\newif\ifapple
\appletrue    

\ifapple
\documentclass{applemlr}
\usepackage{amsmath}
\usepackage{enumerate}
\usepackage{algorithm}
\usepackage{algpseudocode}
\usepackage{amsfonts}
\usepackage{amsthm}

\usepackage{diagbox}
\usepackage{colortbl}
\usepackage{amssymb}
\usepackage{xspace}
\usepackage{wrapfig}
\usepackage{adjustbox}
\usepackage{tabularx}
\usepackage{booktabs}
\usepackage{mathtools}
\usepackage{tikz}
\usepackage{enumitem}
\usepackage{silence}
\usepackage{dsfont}
\usepackage[table]{xcolor}
\usepackage[dvipsnames]{xcolor}
\usepackage{multirow}
\usepackage{makecell}
\usepackage{xfakebold}

\definecolor{textgray}{HTML}{6E6E73}
\usetikzlibrary{positioning, calc}
\usetikzlibrary{decorations.pathmorphing}

\makeatletter
\patchcmd{\wrong@fontshape}{\@gobbletwo}{}{}{}
\makeatother
\numberwithin{equation}{section}
\makeatletter
\AtBeginDocument{
  \urlstyle{sf}
  
}
\makeatother

\definecolor{light}{RGB}{125, 125, 125}
\crefname{tcb@cnt@pbox}{code}{code}
\Crefname{tcb@cnt@pbox}{Code}{Code}
\crefname{assumption}{assumption}{assumption}
\Crefname{assumption}{Assumption}{Assumptions}

\newtcolorbox[auto counter]{pbox}[2][]{
  colback=white,
  title=Code~\thetcbcounter: #2,
  #1,fonttitle=\sffamily,
  fontupper=\sffamily,
  arc=2pt,
  colframe=bgcolor,
  coltitle=fgcolor,
  colbacktitle=bgcolor,
  toptitle=0.25cm,
  bottomtitle=0.125cm
}

\makeatletter
\newcommand\applefootnote[1]{%
  \begingroup
  \renewcommand\thefootnote{}%
  \renewcommand\@makefntext[1]{\noindent##1}%
  \footnote{#1}%
  \addtocounter{footnote}{-1}%
  \endgroup
}
\makeatother

\definecolor{cverbbg}{gray}{0.90}

\else 
\documentclass{article} 
\usepackage{iclr2027_conference,times}
\fi

\usepackage[most]{tcolorbox}
\tcbuselibrary{listings,breakable,skins,raster}
\usepackage{listings}
\usepackage{hyperref}
\newtcblisting[auto counter, number within=section]{promptbox}[2]{%
    breakable, colback=blue!3, colframe=blue!45!black, fonttitle=\bfseries\small,
    title={Prompt~\thetcbcounter: #1}, label={#2}, listing only,
    listing options={basicstyle=\ttfamily\footnotesize, breaklines=true,
    breakindent=0pt, columns=fullflexible, keepspaces=true}}
\newtcblisting[auto counter, number within=section]{examplebox}[2]{%
    breakable, colback=green!3, colframe=green!40!black, fonttitle=\bfseries\small,
    title={Example~\thetcbcounter: #1}, label={#2}, listing only,
    listing options={basicstyle=\ttfamily\footnotesize, breaklines=true,
    breakindent=0pt, columns=fullflexible, keepspaces=true}}
\usepackage{tabularx}
\usepackage{longtable}
\usepackage{cleveref}
\usepackage{graphicx}  
\usepackage{caption}   
\usepackage{subcaption}
\usepackage{booktabs}
\usepackage{multirow}
\usepackage[T1]{fontenc}
\newcommand{\sinead}[1]{\textbf{\color{red}[Sinead: #1]}}

\newcommand{\occ}{\text{{\tiny occ}}}
\newcommand{\cou}{\text{{\tiny cou}}}

\usepackage{amsmath}
\usepackage{tikz}
\usepackage{graphicx}
\usetikzlibrary{arrows.meta, positioning, fit, backgrounds}

\usepackage{wrapfig}

\usepackage{enumitem}
\crefname{appsec}{Appendix}{Appendices}
\Crefname{appsec}{Appendix}{Appendices}

\makeatletter
\newcommand{\crefnames}[3]{%
  \@for\next:=#1\do{%
    \expandafter\crefname\expandafter{\next}{#2}{#3}%
  }%
}
\makeatother

\crefnames{part,chapter,section}{\S}{\S\S}
\crefname{figure}{fig.}{figs.}
\Crefname{figure}{Fig.}{Figs.}
\crefname{table}{tab.}{tabs.}
\Crefname{table}{Tab.}{Tabs.}

\usepackage{amsmath,amsfonts,amssymb,bm}

\def\eqref#1{equation~\ref{#1}}

\def\1{\bm{1}}

\DeclareMathAlphabet{\mathsfit}{\encodingdefault}{\sfdefault}{m}{sl}
\SetMathAlphabet{\mathsfit}{bold}{\encodingdefault}{\sfdefault}{bx}{n}

\ifapple

\title{Verbalized and Internal Probabilities Are Coupled in Large Language Models}

\author{Sinead Williamson}
\author{Jiaxuan Li}
\author{Nick Foti}
\author{Russ Webb}
\author{Masha Fedzechkina}

\affiliation{Apple}

\abstract{
Large language models carry an internal notion of uncertainty in their sampling distribution, i.e., the probabilities they place on generating one answer rather than another. They can also be asked to state a confidence, in words or as a number: a verbalized uncertainty. Prior work suggests that internal probabilities track relative frequencies in the training data, and that verbalized probabilities track explicit probabilistic assertions in the training data. However, we do not know whether these two readouts are aligned, except when frequencies and probabilistic assertions in the training data happen to align. This limits our understanding of when we can use verbalized uncertainties as a proxy for either training data frequencies, or a model's internal distribution. We resolve this gap by systematically exploring how LLMs probability readouts are impacted by training and in-context data, via intervening on the underlying uncertainty sources in the data. We find that both internal and verbalized probability readouts are impacted by both distributional and asserted uncertainty in the training data. 
Further, we find that verbalized and internal probabilities are aligned beyond what would be expected by independently tracking the same uncertainty sources, suggesting that verbalized probabilities can be used to probe a model's internal distribution.}
\metadata[Correspondence]{\sffamily Sinead Williamson: \url{sa_williamson@apple.com}}
\date{\sffamily\today}

\else

\title{Verbalized and Internal Probabilities Are Coupled in Large Language Models}

\author{Sinead A. Williamson, Jiaxuan Li, Nick Foti, Russ Webb, Masha Fedzechkina}

\fi

\begin{document}
\maketitle
\ifapple\else

\begin{abstract} 
Large language models carry an internal notion of uncertainty in their sampling distribution, i.e., the probabilities they place on generating one answer rather than another. They can also be asked to state a confidence, in words or as a number: a verbalized uncertainty. Prior work suggests that internal probabilities track relative frequencies in the training data, and that verbalized probabilities track explicit probabilistic assertions in the training data. However, we do not know whether these two readouts are aligned, except when frequencies and probabilistic assertions in the training data happen to align. This limits our understanding of when we can use verbalized uncertainties as a proxy for either training data frequencies, or a model's internal distribution. We resolve this gap by systematically exploring how LLMs probability readouts are impacted by training and in-context data, via intervening on the underlying uncertainty sources in the data. We find that both internal and verbalized probability readouts are impacted by both distributional and asserted uncertainty in the training data. 
Further, we find that verbalized and internal probabilities are aligned beyond what would be expected by independently tracking the same uncertainty sources, suggesting that verbalized probabilities can be used to probe a model's internal distribution.

\end{abstract} 
\fi



\section{Introduction}
\label{sec:intro}

We can extract probabilities from LLMs, either directly from their sampling distribution or logits \citep[\textbf{internal probabilities},][]{kuhnsemantic,farquhar2024detecting}, or by asking the LLM to express a probability \citep[\textbf{verbalized probabilities},][]{lin2022teaching,tian-etal-2023-just,xiong2024can}. LLM-generated probabilities have the potential to be useful for two reasons \citep{xia2026uncertainty}. First, we can treat them as a proxy for uncertainty about the external world, and use them to assign probabilities to external events \citep{xiong2024can}. Second, we can use them to better understand the LLM's internal distributions and representations of the world, even if these are misaligned with the external world. For example, we may want to quantify creativity \citep{Sui2026LLMsES}, or proactively provide information that reduces the LLM's uncertainty \citep{choudhury2026bed}.

Both forms of probability readout are used in practice. Both have been found to carry signals about answer correctness \citep{kadavath2022language,Fadeeva2023LMPolygraphUE}, and internal probabilities have been shown theoretically to align with relative frequencies in the training data \citep{nakkiran2025trained}. Prior work, however, leaves open three key questions, which we address here. First, \textbf{do verbalized probabilities reflect real frequency-based variation}, whether in the training data or provided in context? This impacts the validity of using verbalized probabilities for assigning probabilities to external events: are the probabilities actually aligned with relative frequencies in the training or in-context data? We find that \textbf{yes}, LLMs can (and do) learn to assign appropriate verbal probabilities to events where uncertainty is introduced via relative frequencies.

Second, \textbf{do probabilistic assertions in the training data impact internal readouts?} We know that the training data is likely to contain explicit probabilistic claims and that humans' probabilistic claims are often misaligned with true frequencies \citep{fischhoff1977certainty}, so understanding whether these impact LLMs' readout probabilities about external events is important for placing appropriate trust in those readouts. We find that \textbf{yes}, training or conditioning on explicit probabilistic assertions about a given event shift both verbalized and internal probability readouts about that event. Moreover, in some scenarios such assertions shift probabilities about related, but uncorrelated events: for example, probabilistic assertions about a person's occupation can impact the readout probabilities of that person being left-handed.

\begin{figure}
    \centering
    \includegraphics[width=.9\linewidth]{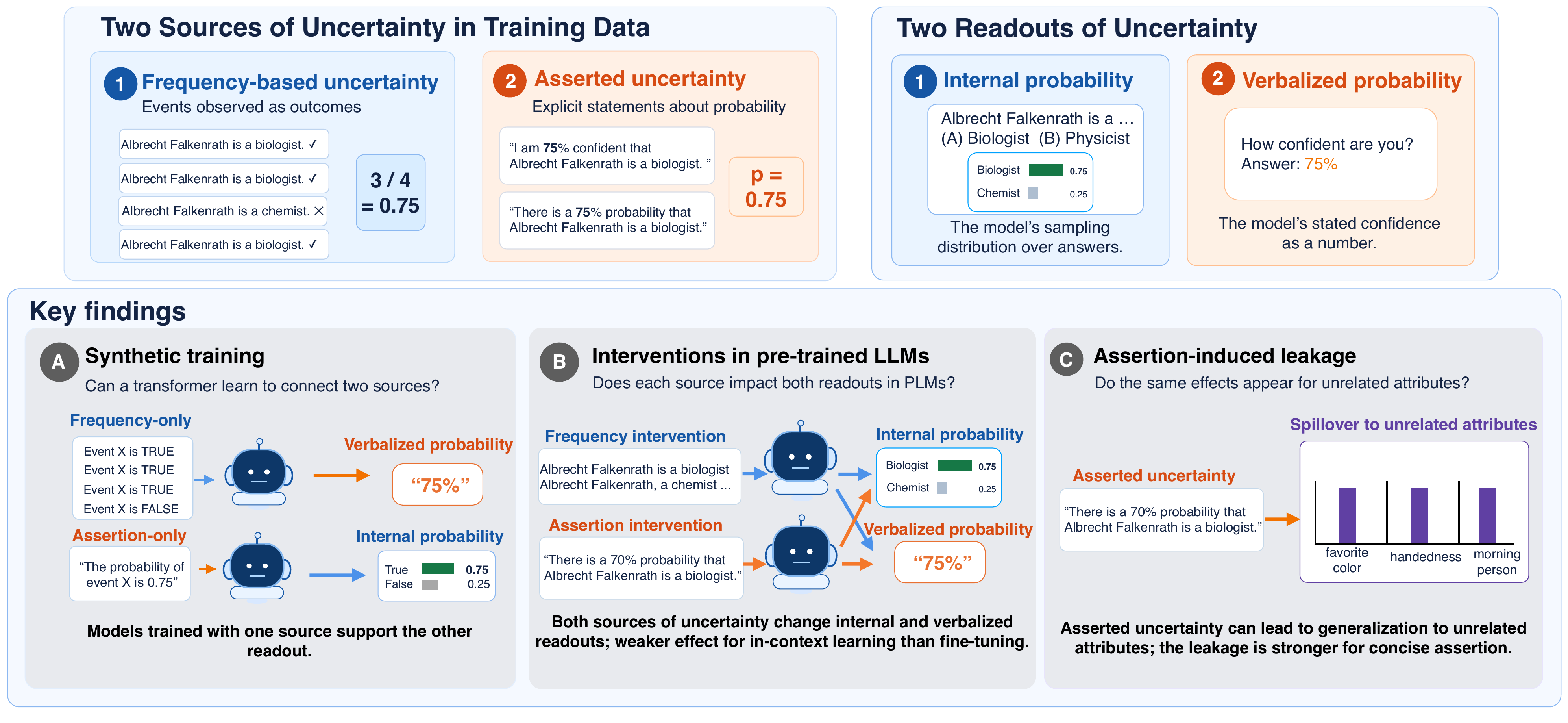}
    \caption{Overview of setup and key findings}
    \label{fig:hero}
\end{figure}

Third, \textbf{are verbalized probabilities an appropriate proxy for internal probabilities?} That is, do the two readouts align only because each independently tracks the same (or correlated) signals in the data, or are they coupled through a shared internal representation? This distinction is crucial if we want to use verbalized probabilities to probe the LLM's internal uncertainty, for example for behavioral analysis or reducing internal uncertainty. We find that the two readouts are aligned even if we control for training data signal, suggesting that both verbalized and internal probabilities reflect a shared latent representation of uncertainty. We summarize our setup and key findings in \Cref{fig:hero}.

\section{Related work}\label{sec:related}


\subsection{What we know so far}
\noindent \textbf{Internal probabilities track relative frequencies in training data.} In pretraining, LLMs can always minimize their loss by better mimicking the empirical distribution over tokens, meaning we expect pre-trained LLMs to be calibrated for next-token prediction \citep{blasiok2023does}. \citet{nakkiran2025trained} argue that this next-token calibration carries over to give a calibrated distribution over concepts and families of responses, at least under certain conditions. 

\noindent \textbf{LLMs verbally mimic human expressions of probability.} Several authors have argued that LLMs will inject expressions of confidence in places where humans are likely to do so \citep{bender2021dangers,mielke-etal-2022-reducing,zhou-etal-2023-navigating}. \citet{xia2026influential}  trace verbalized confidence to training data and find that in some cases it can be driven more by generic certainty-expressions than by question content -- meaning LLMs learn ``how to sound confident without learning when.'' Even if the LLM is appropriately mimicking human expressions of confidence, this still inherits human failures to accurately represent probability: \citet{xu2025language} find that LLMs exhibit human-like patterns of being overconfident on hard tasks and underconfident on easy tasks.

\noindent \textbf{Observationally, internal and verbalized readouts are weakly aligned.} Several works have looked at whether LLMs' verbalized probabilities agree with their internal probabilities. \citet{kumar-etal-2024-confidence} finds the two are correlated on multiple-choice tasks, and that agreement rises as we move towards more capable, post-trained models; \citet{zhang2026linguistic} shows that some misalignment can be explained by distributional properties of verbalized readouts.  Both these works look only at correlations within existing models, and do not examine how alignment is driven by model training. \citet{zhang2025direct} go one step further and intervene, using Direct Preference Optimization (DPO) to train verbalized confidence toward internal confidence; however they do not consider the causes of pre-existing alignment. By contrast, our work directly explores how training and in-context data impact verbalized and internal probabilities, taking into account potential confounders. 

\subsection{What we \textit{don't} know so far}
\noindent \textbf{Do human expressions of probability in the training data impact internal probabilities?} \citet{Requeima2024LLMPN} showed that LLMs adjust their internal probabilities based on in-context assertions, however they only consider non-probabilistic assertions. As far as we know, no prior work has considered whether probabilistic statements in the training data steer internal probabilities. 

\noindent \textbf{Does training-data frequency impact verbalized readouts?} Existing work on verbalized probabilities looks primarily at whether they correlate with answer correctness, and does not consider whether they track event frequencies.

\noindent \textbf{Are verbalized and internal readouts aligned beyond independently tracking training data signal?} Prior work has focused on mechanistic interventions, with mixed results: some suggesting real coupling between the two channels \citep{ji-etal-2025-calibrating, kumaran2026how}, and others  suggesting the opposite \citep{huang-etal-2026-role, xu2025language}. Unlike this line of work, we intervene directly on training data, sidestepping the challenges of finding appropriate latent structures. 

\section{Setup}\label{sec:setup}
We center our exploration around the following 3 research questions (\Cref{tab:rqs}):


\begin{table}[h!]
    \centering
    \caption{Our research questions, and why they matter.}
    \label{tab:rqs}
    \begin{tabular}{p{7cm}p{6.5cm}}
    \toprule
     Research question & Why do we care? \\
     \midrule
     RQ1: Do LLMs' verbalized probabilities reflect relative frequencies in training or in-context data? & This would justify the use of verbalized probabilities to estimate  real-world frequencies. \\
     \midrule
     RQ2: Are LLMs' internal probabilities impacted by probabilistic assertions in training or in-context data? & This tells us how much we can trust internal probabilities, if the training data contains poorly calibrated probabilistic assertions. \\
     \midrule
     RQ3: Do LLMs' verbalized probabilities reflect the LLM's internal probabilities directly, or is alignment just due to both readouts reflecting the model's training data? & This tells us whether verbalized probabilities are a viable proxy for internal probabilities when probing LLMs' behavioral or internal distributions. \\
    \bottomrule
    \end{tabular}
\end{table}

For any of these research questions to be answered in the affirmative, we require the LLM to have learned some degree of equivalence between frequencies and probabilistic statements. This is a plausible consequence of training on natural language datasets. Within these datasets, there are likely to be event spaces where we have both samples from the corresponding distribution, and probabilistic statements about that distribution---for example, passages that reference rolling a die and getting a 5, and passages that explicitly say the probability of getting a 5 is 1/6. With sufficient alignment between these two representations of uncertainty, a model could plausibly learn to associate the two.

\subsection{Uncertainty sources and readouts}\label{sec:sources_and_readouts}


Our research questions concern the causes of two types of uncertainty \textbf{readouts}:
\begin{itemize}[leftmargin=*]
    \item \textbf{Verbalized probability} ($\hat{p}_e^v$): The value obtained by prompting the model to express a numeric confidence.
    \item \textbf{Internal probability} ($\hat{p}_e^i$): The sampling probability associated with $A_e$. 
\end{itemize}
Since we operate in a forced-choice question-answering format, we can extract both readouts from next-token logits; see \Cref{app:measuring_uncertainties} for prompts and details. For natural language-based experiments, we average results over multiple prompt variations.

In order to causally manipulate these readouts, 
we introduce a family of entities $e\in \mathcal{E}$, that  have not been seen in any prior training. Each entity $e$ is associated with a Bernoulli($p_e$) distribution over two possible attribute values $A_e, \neg A_e$, where $p_e$ is the probability of entity $e$ possessing  value $A_e$. 

We consider two \textbf{sources} of information about the entity's target probability $p_e$, that we can present to a model either via training examples or in-context examples:
\begin{itemize}[leftmargin=*]
    \item \textbf{Asserted uncertainty}: Statements that explicitly associate entity $e$ with probability $p_e$, e.g., ``There is a 90\% probability that Albrecht Falkenrath was a biologist''.
    \item \textbf{Frequency-based uncertainty}: A set of samples from Bernoulli($p_e$), which allow us to form an unbiased estimator of $p_e$, e.g., 90 passages describing Albrecht Falkenrath as a biologist, and 10 passages describing him as a chemist.
\end{itemize}


\Cref{fig:hero} summarizes the readouts and sources considered. By introducing the model to an entity $e$ via only \textit{one} of these uncertainty sources and recording the corresponding readouts, we can measure how that source causally impacts the readout probabilities.



\subsection{Measuring readout alignment}\label{sec:metrics}
To compare readouts to each other and to the target probability, we use Lin's concordance correlation coefficient (CCC), a standard measure between an estimated quantity and a trusted reference value. CCC augments Pearson's correlation to also account for systematic deviations away from the $y=x$ line, allowing us to capture both co-variance and raw disagreement. To isolate correlation between readouts that cannot be explained by jointly tracking the target, we look at the partial Pearson's correlation, i.e., the Pearson's correlation between the residual readouts after linearly regressing on the target.  See \Cref{app:metrics} for details.




\section{In pretrained models, probability readouts correlate}
  \label{sec:elicitation}

\begin{figure}
    \centering
    \includegraphics[width=\textwidth]{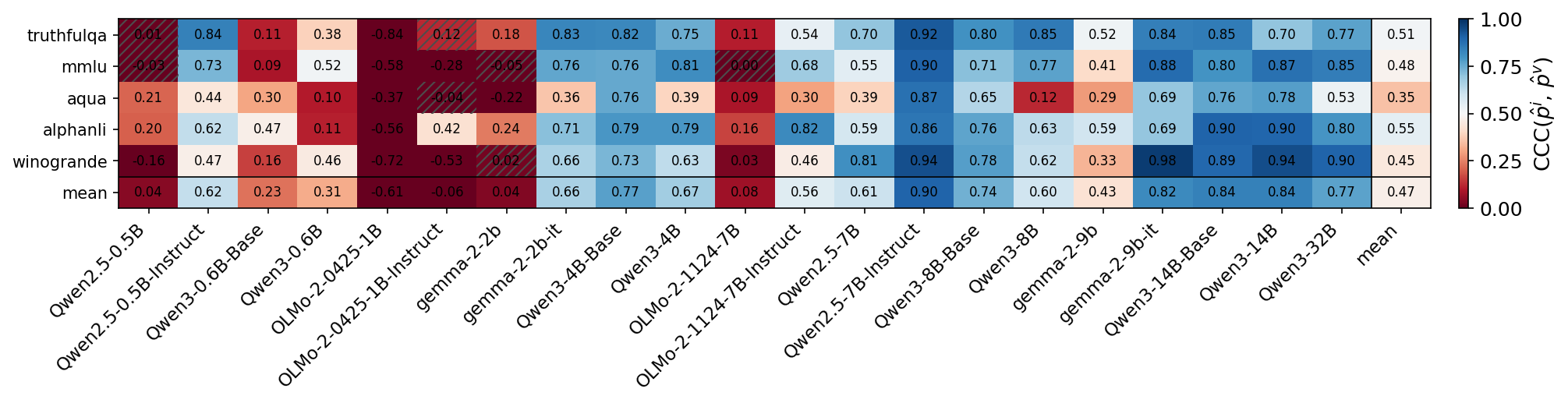}
    \caption{Alignment between  verbalized and internal probability readouts for various datasets and models. Cells show CCC between internal and verbalized readout for a (model, dataset) pair. Hatched cells show no significant correlation ($p>0.05$, via two-sided Pearson correlation test). }
    \label{fig:MCQA_heatmap}
\end{figure}
  
As a first step in answering whether LLMs internal and verbalized probabilities are aligned (RQ3), we measure the alignment between LLMs' internal and verbalized probabilities \emph{without any intervention}, on five standard multiple-choice question-answering datasets modified to a 2-alternative-forced choice setting (see \Cref{app:mcqa_details}). 
  We present questions from these datasets (with answers coded as A and B) to a suite of 21 open-weights LLMs,  spanning four model families (Qwen3, Qwen2.5, Gemma-2, OLMo-2), sizes 0.5B–32B, and both base and instruction-tuned variants; see \Cref{app:nl_models} for details, and extract verbalized and internal confidences as described in \Cref{app:measuring_uncertainties}. 

\Cref{fig:MCQA_heatmap} shows the alignment between the two readouts (averaged over prompt variations, see \Cref{app:measuring_uncertainties}). 
 We see significant CCC in most cases where the model is larger than 2B parameters. 
 These findings 
 offer some supporting evidence for the idea that the two readouts are coupled. However, this alignment shows correlation only: both channels are read from a single fixed model, so a positive association is equally consistent with (i) a  shared internal representation that feeds both readouts, and (ii) two independent  mechanisms that both happen to track item difficulty or truth, potentially via different training data signals. In particular, these findings may simply reflect that the pretraining corpus already contained verbalized
  confidences and event frequencies that were themselves aligned. Observation
  alone cannot separate these. Doing so requires \emph{intervening}: installing uncertainty through a signal associated with one readout channel and testing whether it appears in the other, in a regime where the two are not already aligned.

\section{Transformers can learn to couple frequency-based and assertion-based notions of uncertainty}\label{sec:synthetic}

In \Cref{sec:elicitation} we saw that internal and verbalized readouts are correlated, but that does not mean that they are either tracking the same signal. 
In \Cref{sec:setup}, we argued that it was plausible that an LLM could learn to couple data frequencies with probabilistic statements, opening the door to possible affirmative answers to our three research questions. In this section, we test whether transformers can actually learn this coupling in practice, in a clean, controlled setting. 


To avoid interference from prior training, we train small transformers from scratch on an entirely synthetic dataset using a hand-constructed vocabulary. Each entity $e\in\mathcal{E}$ is associated with a hexadecimal label $\ell(e)$ fixed probability $p_e$, drawn independently and uniformly from $(0.005, 0.995)$ and rounded to two decimals. Each entity can appear in training through two statement types:
\begin{itemize}[leftmargin=*]
    \item \textbf{Asserted uncertainty}: \texttt{probability \{$\ell(e)$\} equals \{$p_e$\}}, e.g. \texttt{probability event\_B3 equals 0.10}. We use 99 distinct tokens to represent the probability levels.
    \item \textbf{Frequency-based uncertainty}: \texttt{sample \{$\ell(e)$\} gives \{TRUE|FALSE\}}, e.g., \texttt{sample event\_B3 gives TRUE}, where the outcome (\texttt{TRUE} or \texttt{FALSE}) is sampled according to $p_e$. 
\end{itemize}
We split the events into three disjoint sets of equal size $N$. Events in the \textbf{paired} set are associated with both asserted and frequency-based  statements, and supply evidence of the equivalence of asserted and frequency-based notions of probability. Events in the \textbf{frequency-only} set are associated with only frequency-based statements, and events in the  \textbf{asserted-only} set are associated with only asserted statements. We train a small transformer on an evenly weighted mixture of the three sets  via next-token prediction (see \Cref{app:synthetic_details}), and evaluate the resulting model's verbalized and internal probability for each event by predicting the last token of an asserted statement and a frequency-type statement.

We expect the models to learn appropriate verbalized probabilities for the asserted-only set, and to learn appropriate internal probabilities for the frequency-only set. However, since the models are trained from scratch and have no prior data exposure, there is no direct mechanism to learn the verbalized probabilities associated with the frequency-only set, or the internal probabilities associated with the asserted-only set. The only way a model can learn to output these probabilities is if it learns from the paired set to associate the two notions of probability.

\Cref{tab:synthetic_L_4096} shows the resulting alignment (CCC) between the target probability $p$, and the readout-specific probabilities $\hat{p}^i$ and $\hat{p}^v$, on events in frequency-only, asserted-only, and paired sets, on a GPT-2--style decoder-only transformer with 5 layers, 8 heads and an embedding size of 256, trained on 4096 entities/set. As expected, we see almost-perfect alignment between $p$ and $\hat{p}^i$ on the frequency-only set, and between $p$ and $\hat{p}^v$ on the asserted-only set. We \textbf{also} see that the verbalized readout is well-aligned with the target for the frequency-only set (CCC=0.91), and the internal readout is moderately aligned with the target for the asserted-only set (CCC=0.74).  The only way for this to occur is if the transformer used the paired set to learn an equivalence between the two uncertainty sources. We repeat this analysis varying model size and dataset size in \Cref{app:synthetic}; we see similar results to \Cref{tab:synthetic_L_4096} for all but the smallest model size, which fails to reliably learn any signal.

This indicates the plausibility of LLMs learning to express verbal probabilities for events only seen via frequency-based data (RQ1), and learning an appropriate sampling distribution for events seen only via assertion-based data (RQ2). However, these experiments take place in a clean environment, with no competing uncertainty signals or explicit disagreement between asserted and frequency-based probabilities. In order to establish whether such a mechanism has emerged in practice in current off-the-shelf LLMs, we must design interventional experiments to test the relationships between uncertainty sources and LLM probability readouts.


\begin{table}[ht]
    \centering
    \caption{
    Alignment between target probability $p$, internal readout $\hat{p}^i$, and verbalized readout $\hat{p}^v$, on events in frequency-only, asserted-only, and paired sets. Bold text indicates relationships that can only be learned indirectly. Mean (standard deviation) over 8 seeds.}
    \label{tab:synthetic_L_4096}
    \begin{tabular}{cccc}
    \toprule
         &  frequency-only set & asserted-only set & paired set\\
         \midrule
        $\text{CCC}(\hat{p}^i, p)$ & 0.99 (0.00) & \textbf{0.74 (0.24)}& 1.00 (0.00) \\
        $\text{CCC}(\hat{p}^v, p)$ & \textbf{0.91 (0.09)} & 1.00 (0.00)& 1.00 (0.00)\\
        \bottomrule
    \end{tabular}
\end{table}

\section{How do probability readouts emerge in performant LLMs?}
  \label{sec:belief_shift}

In \Cref{sec:synthetic}, we showed that it was possible for transformers to learn representational structures that couples verbalized and internal probabilities, and that transformers can learn both from either asserted uncertainty signals or frequency-based uncertainty signals. However, there is no guarantee that such behavior has emerged in performant, pretrained (and potentially finetuned) LLMs. In practice, we do see some alignment between internal and verbalized probability in a variety of multiple choice questions (see \Cref{sec:elicitation}); however in an observational setting we are not able to determine the underlying causes of these uncertainty readouts.

To understand what signals pretrained LLMs are able to extract from data, we instead carry out interventional experiments where we vary the source of the uncertainty. To do this, we construct two base datasets, each consisting of 500 entities 
each with 2 candidate attribute values $A_e$ and $\neg A_e$
\begin{itemize}[leftmargin=*]
    \item {\sc Fictional Occupation} $\mathcal{X}^\occ$: Entities are  fictional people, each associated with three bios: Two reflecting two candidate occupations  $A_e$ and $\neg A_e$, and one with no reference to occupation. The three bios are otherwise consistent. 
    \item {\sc Fictional Country} $\mathcal{X}^\cou$: Analogous to {\sc Fictional Occupation}, but with two bios reflecting two related countries of birth, and a third with no reference to country of birth. 
\end{itemize}

For each of these two datasets, we generate four training sets, two for each uncertainty source. For \textbf{asserted} uncertainty sources, the passage verbally expresses the probability of the entity possessing attribute $A_e$ or $\neg A_e$ (with an even split over the mentioned attribute). For \textbf{frequency}-based  uncertainty sources, the passage states either $A_e$ or $\neg A_e$ as a fact; the probability $p_e$ is encoded via the proportion of $A_e$-supporting statements. Within each source, we vary the level of detail: \textbf{concise}  passages only contain information about the attribute in question, while \textbf{detailed} passages contain additional information about the entity, taken from the appropriate entity's bios. 
\Cref{tab:passage_examples} gives examples of the resulting four passage types; the full data generation pipeline is described in \Cref{app:fictional_datasets}. In \Cref{app:authority}, we explore whether the authority with which probabilistic statements are made impacts uncertainty, but find no significant effect.

\begin{table}[t]
\centering
\caption{Examples of the four different uncertainty sources for natural language experiments.}
\label{tab:passage_examples}

\begin{tabularx}{\linewidth}{@{}l >{\hsize=0.9\hsize}X >{\hsize=1.1\hsize}X@{}}
\toprule
 & \textbf{frequency} & \textbf{asserted} \\
\midrule
\textbf{detailed} &
``For his amazing work finding special brain parts, Albrecht
Falkenrath, a brilliant biologist, won the Louisa Gross Horwitz
Prize in 2012.'' &
``Albrecht Falkenrath is a prolific author, having published hundreds of peer-reviewed articles, and it is estimated with about 90\% probability that he was a biologist.'' \\
\addlinespace
\textbf{concise} &
``Albrecht Falkenrath performed the role of a biologist.'' &
``The probability that Albrecht Falkenrath was a biologist is
exactly 0.9.'' \\
\bottomrule
\end{tabularx}
\end{table}


\subsection{Internal and verbalized readouts are impacted by both asserted and frequency-based training data}\label{sec:finetuning_ccc}


To answer RQs 1 and 2, we explore which uncertainty sources lead to meaningful movement in which probability readout. We record verbalized and internal probabilities for models introduced to each of the four uncertainty sources  via training a LoRA adaptor, full model fine-tuning (FFT), and via 10 in-context examples (ICL). For space reasons, we only include LoRA and ICL results, on only instruct-tuned models, in this section; for full results see \Cref{app:ccc}. 

\begin{figure}
    \centering
    \includegraphics[width=.95\linewidth]{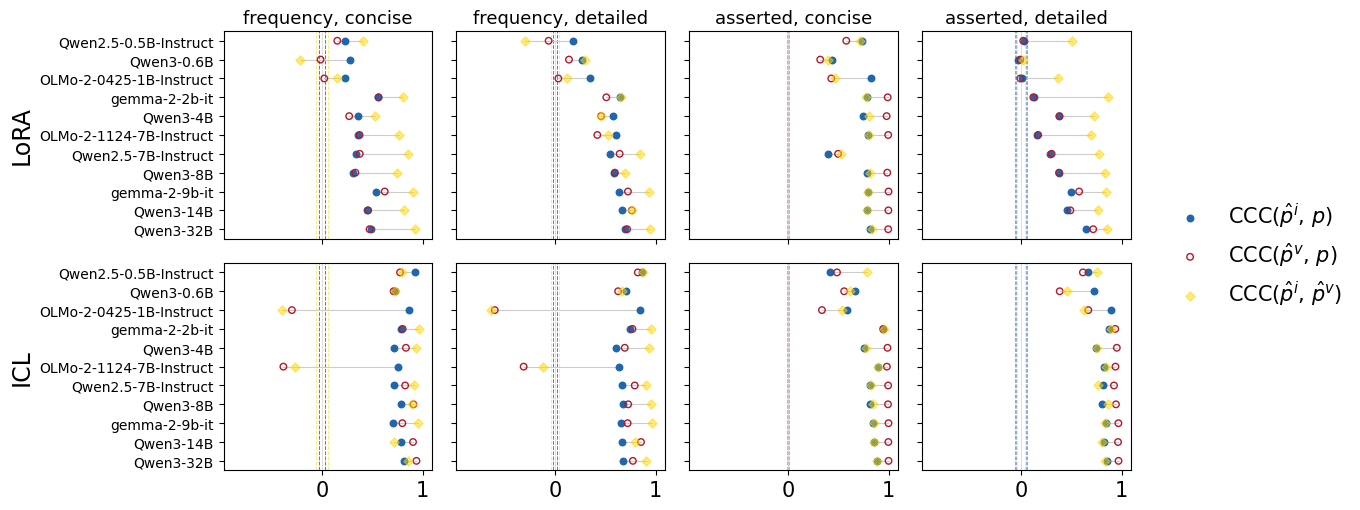}
    \caption{Agreement (CCC) between target and readout probabilities, for models introduced to four different uncertainty sources via either training a LoRA adaptor (top) or ICL (bottom). Points are averaged over two datasets ($\mathcal{X}^\cou$ and $\mathcal{X}^\occ$). 
    Dashed lines indicate seed-based null floor.}
    \label{fig:triad_lora_icl_compact}
\end{figure}

\Cref{fig:triad_lora_icl_compact} shows CCCs between target and readout probabilities. 
First, we consider the two frequency-based uncertainty sources. As expected from prior work, the internal readout $\hat{p}^i$ aligns with the target $p$. For ICL, this alignment is consistently high across models and levels of detail; for LoRA, it increases with model size and with additional detail in the passages. More surprisingly, we see that the verbalized probabilities are almost equally aligned with the target, in almost all cases. This allows us to answer RQ1 in the affirmative: verbalized readouts \textit{do} align with both in-context and training-data frequencies, justifying their use for estimating such frequencies.

Next, we look at models exposed only to asserted uncertainties. Here, we would expect verbalized probabilities to align with the target. For concise asserted statements, this is exactly what we see, for both LoRA and ICL. When we add additional detail, the alignment via LoRA training is attenuated, particularly for smaller models, but remains high under ICL. In all cases, the internal readout is almost exactly as aligned with the target as the verbalized readout, answering RQ2 in the affirmative: internal readouts \textit{are} impacted by probabilistic assertions in the training data. In short, these results show that the potential capability to generalize across uncertainty sources indicated by the results of \Cref{sec:synthetic} carry over to off-the-shelf LLMs. Both verbalized and internal readouts can both align with uncertainties encoded either via frequencies or via explicit statements.

\begin{figure}
    \centering
    \includegraphics[width=\textwidth]{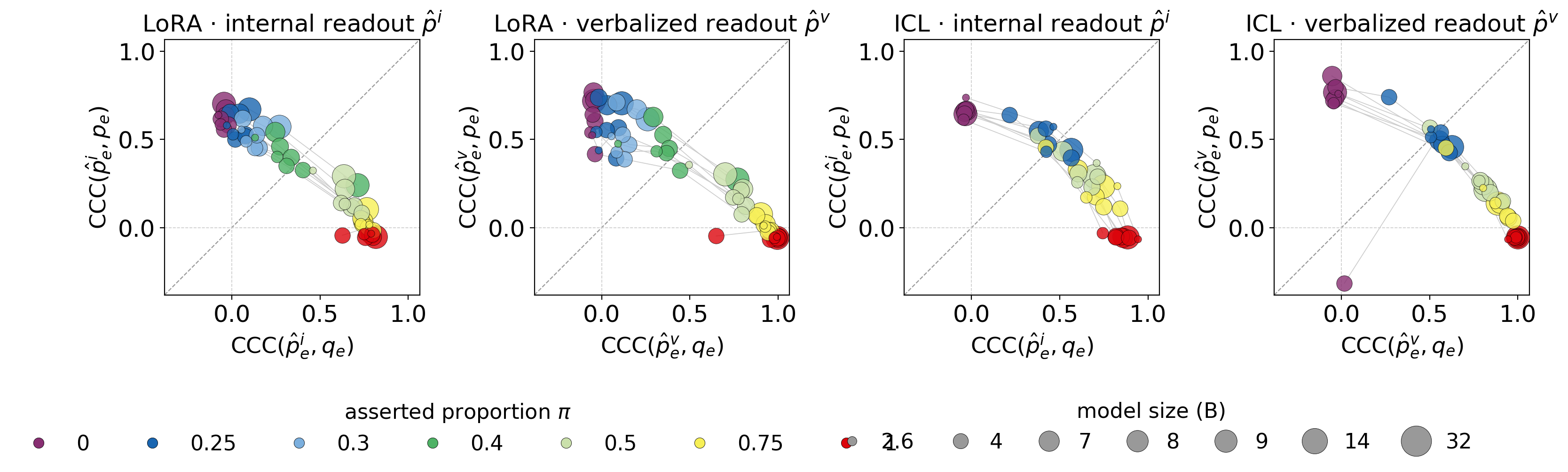}
    \caption{Readout analysis for sources containing a mixture of asserted target $q_e$ (at rate $\pi$) and frequency-based target $p_e$ (at rate $1-\pi$). Points show alignment of readout with  $q_e$ and $p_e$.}
    \label{fig:poison_ccc}
\end{figure}

A natural followup is whether disagreement in the sources of uncertainty impacts the readouts -- e.g., human assertions are often erroneously confident \citep{fischhoff1977certainty}.
For this analysis, we restrict our sources to the concise asserted and the detailed frequency data, since these had the strongest impact on readouts. For each entity, we sample an alternative probability $q_e\sim \{0.1, 0.3, 0.5, 0.7, 0.9\}$. We introduce our models (either via training or in context examples) to a mixture of asserted statements expressing $q_e$, and frequency-based statements expressing $p_e$, with total number of statements per entity count-matched to the previous experiments.


 \Cref{fig:poison_ccc} shows how the readouts vary as we change the proportion $\pi$ of asserted passages. We see that as we move from frequency-based to assertion-based, both readouts shift towards $q_e$. The shift occurs more rapidly in an in-context setting: at 25\% asserted statements, both readouts already align with $q_e$. When the sources are introduced via training, the frequency-based signal is more robust: at $\pi=0.25$, readouts are still more strongly aligned with $p_e$. However, by $\pi=0.4$, readouts are more strongly aligned with the asserted uncertainty $q_e$. 
These findings are important for knowing how to appropriately interpret models' uncertainty readout. Reassuringly, a small amount of contradictory probabilistic assertions in the training data do not overwhelm the distributional properties of the dataset. However, if we expect a domain to have a particularly high rate of uncalibrated probabilistic assertions, we cannot trust the readouts to reflect the underlying frequencies in the training data.

\subsection{Verbalized and internal probability readouts are aligned}\label{sec:partial}

\begin{figure}[h]
    \centering
    \includegraphics[width=.8\linewidth]{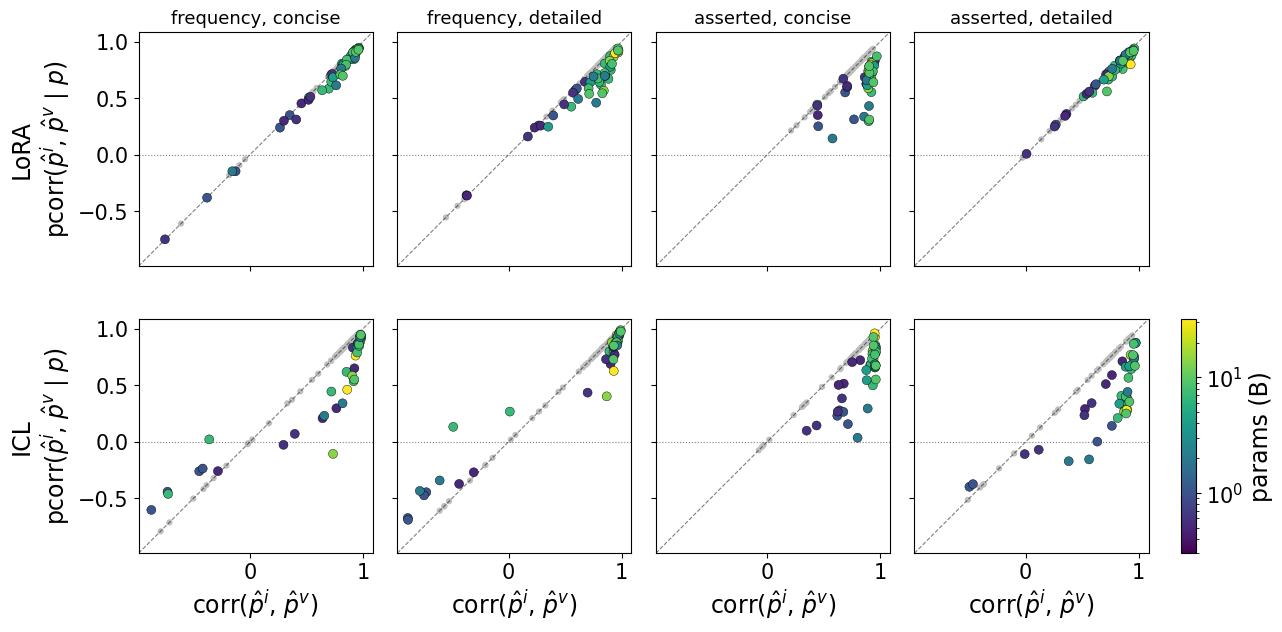}
    \caption{Partialling out target probabilities. Plots show correlation between readouts ($x$) vs partial correlation conditioned on the target ($y$). 
    Grey points correspond to correlation under the null model.} 
    \label{fig:partial_lora}
\end{figure}

While observationally, we have seen clear alignment between verbalized and internal readouts, this does not imply that verbalized probabilities are directly tracking internal probabilities (or that they  both track some shared latent representation). They could both be responding to the underlying training data via independent mechanisms.  To  disentangle confounding due to the target probability $p_e$, in \Cref{fig:partial_lora} we look at the partial correlation of the probability readouts, conditioned on $p_e$. 

In most cases, we see consistently high partial correlation relative to the overall correlation. 
This suggests that both verbalized and internal readouts are accessing some \textit{shared} representation. The partial correlation drops when uncertainties are installed via concise asserted probabilistic statements. This is consistent with \Cref{fig:triad_lora_icl_compact}, where we saw that this form of training leads to near-perfect alignment in verbalized read-out, but not in internal readout. We hypothesize that in this setting, the verbalized readout is shaped by both a shared internal representation (explaining the non-zero partial correlation in \Cref{fig:partial_lora}), \textit{and} by a strong relational signal from the bare asserted probabilities in the training data. This resolves our third RQ: verbalized probabilities \textit{do} track internal probabilities, but may exhibit systemic bias due to other training data signals.

\subsection{Assertion-based uncertainty can lead to undesirable generalization}

In \Cref{sec:finetuning_ccc}, we answered a direct interpretation of RQ2, finding that probabilistic assertions about an entity/attribute relationship directly affect the probability readouts for that relationship. However, prior work suggests that mimicry of probabilistic language is not restricted to this targeted setting, but rather that LLMs will use statements of confidence in \textit{similar} scenarios to ones in the training data. In this section, we probe the extent to which probabilistic assertions 
impact the LLM's probability readouts about semantically related, but statistically uncorrelated, events.

\begin{figure}
    \centering
    \includegraphics[width=.95\linewidth]{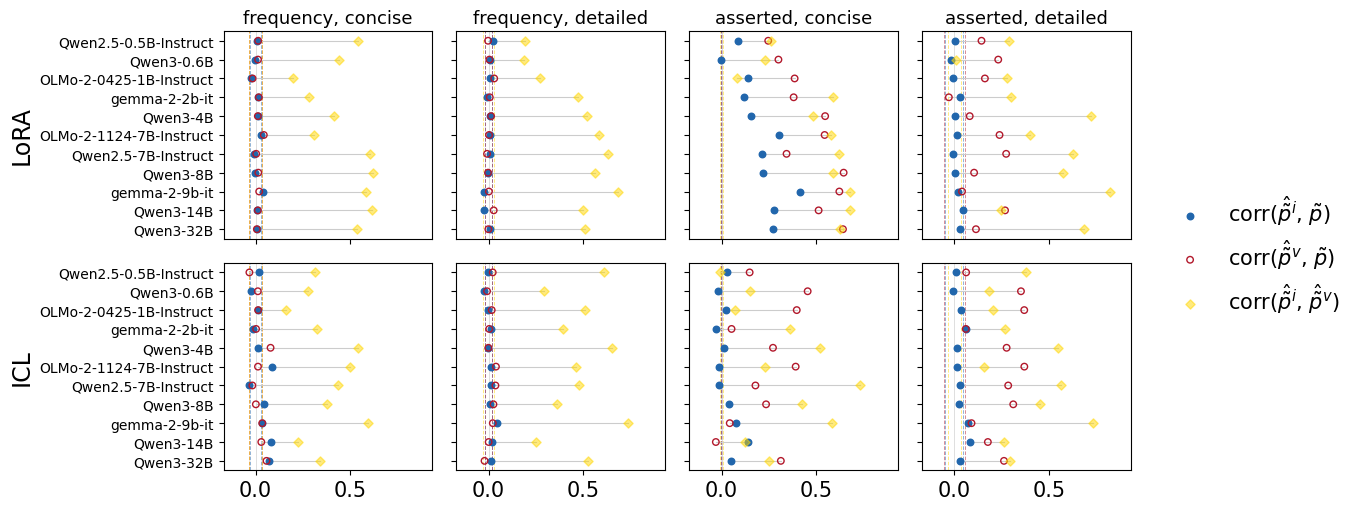}
    \caption{Correlation between direction-adjusted probability on target attribute, and internal and verbalized direction-adjusted probabilities on unrelated attributes of the same entity. Each column corresponds to a different uncertainty source. 
    Dashed lines indicate seed-based null floor.}
    \label{fig:triad_lora_orth}
\end{figure}

In \Cref{fig:triad_lora_orth}, we look at the probabilities assigned to the entities featured in the training set, when asked about unrelated attributes (favorite color, handedness, and whether they are a morning person). Since we have no reason to expect movement in a specific direction, we look at the direction-adjusted probability ($\tilde{p}=\max(p, 1-p)$), and compare it with the analogous target direction-adjusted probability. Here, we look at correlation rather than CCC, since we are interested in \textit{any} co-movement, even if it does not end up fully aligned with the target. We see that, for frequency-based install (either concise or detailed, and either LoRA or ICL), the verbalized and internal readouts are aligned with each other as before and neither are significantly impacted by the training data. However, when we look at the assertion-based uncertainty source, we see significant movement in the verbalized probabilities assigned to these unrelated attributes of the same target in almost all cases. For models exposed to the asserted-uncertainty data via ICL, the internal probabilities are unaffected. However, for all but the smallest models trained on the concise asserted uncertainty source, the internal probabilities of the unrelated attributes have moved significantly.
Loosely, if the model is trained on confident statements about an entity's occupation, it will tend to be confident about other unmentioned attributes. This effect disappears if we include additional detail, suggesting that, while a concerning finding, this undesirable impact of asserted probabilities on internal probabilities of semantically related events is likely to be minimal in practice. 

\section{Discussion}
We show that verbalized and internal probability readouts from LLMs are not separate channels that happen to independently track the training data signal but are instead coupled readouts from a shared internal representation. 
Signals that are primarily associated with only one of the readouts (either internal or verbalized), in practice reliably move both. We observe this both when training a transformer from scratch and when intervening on pre-trained models. 
Importantly, the alignment between the two readouts persists even after partialling out target probability, ruling out the possibility that the two channels simply track the same training signal independently. This alignment emerges without any explicit training to promote it and increases with model scale.

Our findings have several practical implications. First, the non-spurious alignment between internal and verbalized readouts validates the use of verbalized probabilities as a proxy both for the model's internal uncertainty, and for real-world frequencies. Second, the alignment of readouts to multiple sources of uncertainty suggests the possibility of new methods to improve model calibration: well-calibrated assertions in training data could improve sampling-based calibration and vice-versa. Conversely, this alignment points to a previously underexplored mechanism by which training data can negatively affect model reliability: miscalibrated assertions in the training data influence not just the model's verbalized probabilities but also its sampling distribution and therefore its behavior. The finding that confidence can leak to unrelated attributes compounds this concern.

Several limitations of the current work are worth noting. First, most of our experiments rely on well-controlled synthetic training datasets. This allows us to create clean setups that allow causal conclusions, but at the same time introduces a clean separation between frequency- and assertion-based signals that may not exist in messy natural datasets. Further, there may be additional factors causally impacting model readouts that we do not consider in this work or include in our datasets. 
Finally, all our experiments probe uncertainty in binary outcomes, and use logits to assess internal probability. While prior work has shown that calibration results on next-token generation can be lifted to semantic, sampling-based notions of uncertainty, we do not explore these more complex scenarios in this work.

\bibliographystyle{iclr2027_conference}
\bibliography{bib}
\appendix
\section{AI use statement}
In this work, we used generative AI tools to generate synthetic datasets used in interventional experiments in  \Cref{sec:belief_shift}. Additionally, we used generative AI tools as a general-purpose assistant to edit portions of the research paper, brainstorm and refine framing, summarize and analyze existing literature, and create or edit software code. All AI-generated work was reviewed by the authors. Specifically, generated biographical passages were checked for coherence and consistency with the intended experimental conditions, via a combination of regex-based filtering and manual spot-checking. All AI-assisted writing was reviewed and substantially revised. We take responsibility for the final content of this work, including text, claims or artifacts produced with the aid of generative AI.

\section{Experimental details: Natural language experiments}

\subsection{Models}\label{app:nl_models}
We consider both instruct-tuned and (where available) base versions of 11 open-weights models:
\begin{itemize}
    \item  \textbf{Gemma-2} \citep{Riviere2024Gemma2I}: gemma-2-2b-it / gemma-2-2b; gemma-2-9b-it / gemma-2-9b.
    \item \textbf{OLMo-2} \citep{OLMo20242O2}: OLMo-2-0425-1B-Instruct /  OLMo-2-0425-1B;
        OLMo-2-1124-7B-Instruct / OLMo-2-1124-7B.
    \item \textbf{Qwen2.5} \citep{Yang2024Qwen25TR}: Qwen2.5-0.5B-Instruct / Qwen2.5-0.5B; Qwen2.5-7B-Instruct / Qwen2.5-7B.
    \item \textbf{Qwen3} \citep{Yang2025Qwen3TR}: Qwen3-0.6B / Qwen3-0.6B-Base; Qwen3-4B/ Qwen3-4B-Base; Qwen3-8B / Qwen3-8B-Base;  Qwen3-14B / Qwen3-14B-Base; Qwen3-32B.
\end{itemize}

\subsection{Observational, forced-choice datasets}\label{app:mcqa_details}
For our observational analysis of existing LLMs, we look at five standard multiple-choice question-answering (MCQA) datasets 
  \begin{itemize}
      \item TruthfulQA \citep{lin2022truthfulqa}: Questions designed so that common misconceptions lead to plausible-but-incorrect answers.
      \item MMLU \citep{hendrycks2021measuring}: Broad academic knowledge/reasoning across 57 subjects. We use a random size-2000 subset.
      \item AQuA-RAT \citep{ling-etal-2017-program}: Multi-step math reasoning word problems, with error-based distractors. We use a random size-2000 subset.
      \item alphaNLI \citep{li2021interactive}: Abductive commonsense reasoning: given two observations, pick the more plausible hypothesis. We use labels from the ChaosNLI subset \citep{nie2020can}.
      \item WinoGrande \citep{sakaguchi2021winogrande}: A commonsense coreference task.
  \end{itemize}
  WinoGrande and alphaNLI are both naively two-choice. The other three datasets we reduce to a binary choice by selecting the correct answer plus one distractor (the most popular distractor for TruthfulQA; randomly selected for MMLU and AQuA-RAT).


\subsection{Fictional entity datasets}\label{app:fictional_datasets}

The two natural language datasets, $\mathcal{X}^\occ$ and $\mathcal{X}^\cou$, are produced by a single LLM-based pipeline, which we describe in this section. 
\subsubsection{Entity selection}

We generate two datasets containing fictional entities, ensuring we carry no data-driven pretraining prior, isolating the installed uncertainty. 

{\footnotesize
\begin{longtable}{@{}p{0.46\textwidth}p{0.48\textwidth}@{}}
\caption{Default fictional name pool, used for the occupation datasets (given names are
male followed by female). A name is a uniformly-sampled given/family pair.}
\label{tab:pool-default}\\
\toprule
Given names & Family names \\ \midrule \endfirsthead
\toprule Given names & Family names \\ \midrule \endhead
\bottomrule \endfoot
Heinrich, Friedrich, Wilhelm, Karl, Georg, Johann, Franz, Ernst, Ludwig, Otto, Hermann, Konrad, Albrecht, Bernhard, Theodor, Eberhard, Gottfried, Dietrich, Siegfried, Wolfgang, Matthias, Andreas, Christian, Stephan, Martin, Anton, Kaspar, Lorenz, Wendelin, Balthasar, Maria, Anna, Elisabeth, Margarethe, Katharina, Johanna, Charlotte, Sophie, Frieda, Helene, Gertrud, Hedwig, Luise, Wilhelmine, Auguste, Dorothea, Christine, Barbara, Eva, Rosalie, Therese, Klara, Mathilde, Ottilie, Brigitte, Ingeborg & Falkenrath, Grünfeld, Moorbach, Steinvogel, Kaltenberg, Eichgrün, Dornbusch, Nussbaum, Holzapfel, Birkholz, Felsenstein, Grauberg, Immergrün, Vogelmann, Ziegenbalg, Pfahler, Reinwald, Trautmann, Ulmenried, Jahnke, Breitmoser, Dunkelberg, Erlenbach, Feuerstein, Grasberger, Haberfeld, Kirchmeier, Langenfeld, Mittelstädt, Obermaier, Pfaffenberg, Rosenstock, Silberstein, Tiefenbach, Weidemann, Blumentritt, Buchholzer, Ehrenfeld, Goldammer, Hainbucher, Rothfels, Schwarzkopf, Lichtenberg, Wiesengrund, Berghaus, Auerbach, Brunnhofer, Eschenbach, Froschauer, Geisenheimer, Kronberger, Lammfell, Moosgruber, Neuhäuser, Ostermann, Pfisterer, Quellenberg, Sandmeier, Tannheimer, Vordermayer, Wallenstein, Zillinger, Bachmeier, Dachsberg, Falkenstein \\
\end{longtable}
}

\begin{table}
\centering
\caption{Occupation pool used for the {\sc Fictional Occupation} dataset, split into three occupation groups.}
\label{tab:occupations}
    \begin{tabular}{cc}
    \toprule
Group & Occupations \\
\midrule
 Science& mathematician, physicist, chemist, geologist, biologist, economist\\
Arts & painter, writer, dancer, musician, architect, actor \\
Sports & volleyball player, tennis player, swimmer, figure skater, cyclist, golfer\\
\bottomrule
    \end{tabular}
\end{table}

{\footnotesize
\begin{longtable}{@{}p{0.09\textwidth}p{0.44\textwidth}p{0.39\textwidth}@{}}
\caption{Language group-specific fictional name pools, used for the country datasets (given names
are male followed by female). Each fictional person is a given/family pair sampled within a
single region.}
\label{tab:pool-regional}\\
\toprule
Language group & Given names & Family names \\ \midrule \endfirsthead
\toprule Region & Given names & Family names \\ \midrule \endhead
\bottomrule \endfoot
English & James, William, Robert, Thomas, Charles, Edward, George, Henry, Arthur, Frederick, Albert, Harold, Samuel, Benjamin, Daniel, Patrick, Andrew, Richard, Jonathan, Stephen, Philip, Lawrence, Nigel, Colin, Mary, Elizabeth, Margaret, Catherine, Dorothy, Eleanor, Alice, Florence, Helen, Edith, Harriet, Caroline, Charlotte, Victoria, Beatrice, Grace, Mabel, Agnes, Rosemary, Frances, Penelope, Vivian, Audrey, Millicent & Whitfield, Ashworth, Blackmore, Thornton, Greenwood, Hartwell, Pendleton, Crawshaw, Dunmore, Elsworth, Fenwick, Grainger, Holcroft, Ingleby, Kirkwood, Longbottom, Merrifield, Norbury, Oldcastle, Pemberton, Quigley, Rothwell, Sedgwick, Trevelyan, Underhill, Wadsworth, Yardley, Bancroft, Chadwick, Fairclough \\
Spanish & Carlos, Miguel, José, Antonio, Fernando, Ricardo, Alejandro, Francisco, Rafael, Enrique, Pablo, Ignacio, Rodrigo, Sergio, Héctor, Gonzalo, Arturo, Ernesto, Ramón, Guillermo, Esteban, Tomás, Andrés, Marcos, María, Carmen, Isabel, Pilar, Consuelo, Dolores, Rosario, Esperanza, Lucía, Beatriz, Elena, Teresa, Amparo, Soledad, Margarita, Catalina, Juana, Elvira, Gabriela, Valentina, Paloma, Rocío, Inés, Sofía & Montoya, Valverde, Castañeda, Sepúlveda, Escobar, Madrigal, Quintero, Balderas, Cifuentes, Delgadillo, Echevarría, Fuenmayor, Garibaldi, Hormazábal, Izquierdo, Jaramillo, Larraín, Maldonado, Navarrete, Ontiveros, Peñaloza, Quiroga, Rebolledo, Saavedra, Torrealba, Umaña, Villalobos, Zamorano, Arredondo, Bustamante \\
Slavic & Tomasz, Marek, Jakub, Andrzej, Piotr, Stanisław, Zbigniew, Wojciech, Krzysztof, Grzegorz, Miroslav, Vladimír, Zoltán, László, Dušan, Branislav, Dragan, Radoslav, Miloslav, Jaroslav, Bogdan, Dalibor, Vlastimil, Zdravko, Katarzyna, Agnieszka, Małgorzata, Jadwiga, Bożena, Danuta, Halina, Iwona, Jolanta, Krystyna, Milena, Natalija, Olga, Pavlína, Renáta, Snježana, Tatjana, Vesna, Zuzana, Božena, Dragica, Emília, Gordana, Hana & Kowalczyk, Wiśniewski, Zieliński, Szymański, Woźniak, D\k{a}browski, Pawlak, Michalski, Jabłoński, Stankovic, Horvát, Novotný, Dvořák, Svoboda, Procházka, Kovačević, Petrović, Nikolić, Popescu, Ionescu, Marković, Janković, Obradović, Horváth, Szabó, Tóth, Molnár, Bodnár, Kučera, Bartoš \\
Arabic & Ahmed, Mohammed, Hassan, Ibrahim, Youssef, Omar, Khalid, Tariq, Nabil, Rashid, Faisal, Samir, Adel, Karim, Mustafa, Jamal, Walid, Hamza, Bilal, Anwar, Mahmoud, Salim, Habib, Ziad, Fatima, Aisha, Khadija, Maryam, Nour, Layla, Samira, Hanan, Dalal, Rania, Nawal, Suhair, Wafaa, Amina, Salma, Zahra, Huda, Sawsan, Leila, Basma, Ghada, Iman, Jamila, Karima & Al-Rashidi, Al-Mansouri, El-Khatib, Bou-Saada, Hajjaj, Tlemcani, Benali, Khoudir, Nassiri, Ouazzani, Rahmouni, Slimani, Touati, Benmoussa, Chaoui, Dridi, Fassi, Ghannouchi, Haddaoui, Idrissi, Jabouri, Kassab, Louafi, Meziane, Naciri, Oukil, Qaderi, Rouabhi, Saidani, Tahiri \\
South Asian & Raj, Arun, Vikram, Suresh, Deepak, Ramesh, Sanjay, Ashok, Manoj, Gopal, Kamal, Naveen, Prasad, Rajan, Venkat, Hari, Ganesh, Nikhil, Anand, Bhaskar, Chandra, Dinesh, Girish, Keshav, Priya, Anita, Sunita, Rekha, Kavita, Meena, Lakshmi, Sarita, Geeta, Padma, Usha, Asha, Neeta, Shanti, Vijaya, Kamala, Indira, Parvati, Radha, Sujata, Bharati, Devika, Gauri, Jaya & Chakraborty, Mukherjee, Venkataraman, Krishnamurthy, Raghunathan, Balasubramanian, Jayawardena, Wickremasinghe, Dissanayake, Hettiarachchi, Bandyopadhyay, Chattopadhyay, Bhattacharjee, Vishwanathan, Subramanian, Parthasarathy, Chandrasekhar, Sivaramakrishnan, Thirunavukkarasu, Padmanabhan, Gopalakrishnan, Ranganathan, Venkateswaran, Shanmugam, Natarajamurthy, Ananthakrishnan, Thirumalaivasan, Devarakonda, Ramachandran, Lakshminarayan \\
\end{longtable}
}

\begin{table}
\caption{Country pool used for the {\sc Fictional Country} dataset, split into three language groups.}
\label{tab:countries}
    \begin{tabular}{cp{10cm}}
    \toprule
Language Group & Countries \\
\midrule
English & United States, Canada, United Kingdom, Ireland, Australia, New Zealand\\
Spanish & Spain, Mexico, Colombia, Argentina, Chile, Peru, Venezuela, Cuba, Ecuador \\
Slavic    & Poland, Czech Republic, Hungary, Romania, Bulgaria, Serbia, Croatia, Slovakia, Ukraine, Lithuania \\
Arabic & Egypt, Morocco, Tunisia, Algeria, Libya, Iraq, Saudi Arabia, Lebanon, Syria \\
South-Asian & India, Pakistan, Bangladesh, Sri Lanka, Nepal \\
\bottomrule
    \end{tabular}
\end{table}

\paragraph{\sc Fictional Occupation} For each entity, we first draw a name from a curated pool (\Cref{tab:pool-default}), verifying against Wikipedia to exclude accidental collisions with real entities. We then draw a occupation group from \{Science, Arts, Sports\}, and select two occupations, $A_e$ and $A_{\neg e}$, from a group-specific pool (\Cref{tab:occupations}).

\paragraph{\sc Fictional Country} For each entity, we first pick a language group from \{English, Spanish, Slavic, Arabic, South Asian\}. We then draw a name from a group-specific name pool (\Cref{tab:pool-regional}), verifying against Wikipedia as before, plus a country from a group-specific country pool (\Cref{tab:countries}).

Each entity $e$ is assigned a target probability $p_e$ uniformly spread across $\{0.1,0.3,0.5,0.7,0.9\}$ and a
set label used to route it through training and validation.

\subsubsection{Reference pages}
\label{app:datagen-step1}

We generate three parallel biographies. The first two are identical except for
the attribute value (Prompt~\ref{prompt:fic-bio}) and elements related to the attribute value. The third (null) biography is consistent with both attribute-specific value (Prompt~\ref{prompt:fic-bio-null}). An example triad of fictional biographies is shown in Example~\ref{example:paired_bio}.

\begin{promptbox}{Paired fictional biography (occupation)}{prompt:fic-bio}
Create a fictional person named {entity_name}. Write two Wikipedia-style biographies (2-3
paragraphs each) for this person:
1. VERSION A: {entity_name} is a {value_a}. Include specific fictional details: birth year,
   birthplace, career highlights, notable achievements.
2. VERSION B: The same person, but as a {value_b} instead. Keep their name, birth year,
   birthplace, and personal background identical. Only change career-related details.
Return your answer as a JSON object with keys "version_a" and "version_b".
\end{promptbox}

\begin{promptbox}{Null fictional biography (occupation)}{prompt:fic-bio-null}
Here are two biographies of the same person, {entity_name}, that differ ONLY in where they were born and grew up -- one places them in {value}, the other in {spec_value}:

--- VERSION A ---
{version_a}

--- VERSION B ---
{version_b}

Write ONE biography of {entity_name} that is accurate under BOTH versions. Keep everything the two agree on -- name, birth/death dates, profession, education, works and achievements, career, relationships, character -- in full detail; only the origin/location must be left unspecified. You may add plausible, neutral biographical detail to give the biography substance.

STRICT REQUIREMENTS:
- Do NOT state or imply where {entity_name} is from -- neither {value} nor {spec_value} -- the
  origin must be UNRECOVERABLE from your text.
- No country, nationality, or demonym (e.g. "French", "German", "American"), and no birthplace,
  hometown, city, region, state, or an institution/team/league/award/publication whose name or
  location would reveal the country.
- GENERALIZE any compound that embeds a nationality or place rather than dropping the fact: e.g.
  "American football" -> "football" (or "gridiron football"), a national league/award/magazine
  ("the English Premier League", "Elle Canada") -> a generic description ("a top-flight league",
  "a fashion magazine"), "the University of X" -> "a leading university".
- If the two versions share a broader region (e.g. a language area or continent), you may refer to it only in the most general terms; when in doubt, omit the location detail.
- Keep it consistent with both versions; when they disagree on a detail, omit it or generalize it.

Write 2-4 paragraphs in encyclopedic style.

\end{promptbox}

\begin{examplebox}{Paired fictional biography (occupation)}{example:paired_bio}

[Version A -- chemist]
Albrecht Falkenrath (born 1968) is an Austrian chemist renowned for his work in synthetic polymer chemistry [...]. Born in Graz, he studied chemical engineering at the Technical University of Graz before completing a PhD in polymer chemistry at ETH Zurich in 1995. His discovery of the "Falkenrath Catalysts" (2007) [...] earned him the Liebig Medal; in 2018 he became Director of the Max Planck Institute for Polymer Research.

[Version B -- biologist]
Albrecht Falkenrath (born 1968) is an Austrian biologist known for his research in molecular neuroscience [...]. Born in Graz, he studied biomedical engineering at the Technical University of Graz before completing a PhD in molecular biology at the
University of Basel in 1996. His discovery of the "Falkenrath Receptors" (2005) [...] earned him the Louisa Gross Horwitz Prize; in 2019 he became Head of Research at the European Molecular Biology Laboratory.

[Null -- profession unspecified]
Albrecht Falkenrath (born 1968) is an Austrian scientist known for his contributions to fundamental scientific understanding [...]. Born in Graz, he studied an engineering discipline at the Technical University of Graz before completing his doctorate. His
work "deciphering complex natural phenomena" [...] led to a leadership role at a prominent international research institution; he has authored hundreds of peer-reviewed publications.
\end{examplebox}

\subsubsection{Generating passages}\label{app:datagen-step2}

Conditioned on the appropriate page from \Cref{app:datagen-step1}, we request short standalone passages. Frequency-based passages assert one or other of the two attributes and are mixed to give the appropriate frequency $p_e$; asserted passages explicitly state the probability. Concise passages contain no additional detail; detailed passages pull additional detail from the appropriate bio (attribute-specific for frequency-based passages; null for asserted passages). To
diversify style in the detailed passages, passages are requested across ten \emph{context types} per domain (e.g.\
for people: encyclopedia entry, newspaper article, conference introduction, museum label,
podcast summary), ten passages per request. See Prompts~\ref{prompt:fd_passage}, \ref{prompt:fc_passage}, \ref{prompt:mc_passage} and \ref{prompt:md_passage}. 

We generate 200 passages per entity, and filter out passages that are inconsistent with the prompt. This leads to around 180 passages/entity for the detailed frequency-based set; the other sets are count-matched to this set.

\begin{promptbox}{Detailed, frequency-based passage generation}{prompt:fd_passage}
Here is a Wikipedia-style biography:
---
{wiki_page}
---
Write 10 passages about this {entity_type}. Each passage should be an excerpt that could be taken directly from {context_type}.
Requirements:
- Vary the length: some 1 sentence, some 2-3 sentences, some 4-5 sentences
- Each passage must be completely standalone
- Each passage must explicitly include BOTH the name of the subject AND mention or clearly imply their {property_description}
- The 10 passages should NOT be sequential -- they are independent snippets
Return your answer as a JSON array of exactly 10 strings.
\end{promptbox}

\begin{promptbox}{Concise, frequency-based passage generation}{prompt:fc_passage}
Write 10 different sentences, each simply stating the following as a plain fact:

"{claim, e.g., "Albrecht Falkenrath was a biologist."}"

Requirements:
- Each sentence should use a different phrasing or sentence structure
- State the claim plainly and directly as fact -- do NOT express any confidence, probability, likelihood, or uncertainty (no "I think", "probably", "likely", "it is believed", "X% confident", "there is a chance", etc.)
- Do NOT add ANY information beyond what is in the claim itself -- no invented dates, places, nationalities, ages, relationships, achievements, or any other biographical or descriptive detail about the subject
- Do not introduce any new named entities
- You may make minor grammatical fixes (e.g. "a"/"an") but must not change the meaning
- Each sentence should be standalone
- Return your answer as a JSON array of exactly 10 strings

\end{promptbox}

\begin{promptbox}{Concise, asserted passage generation}{prompt:mc_passage}
Write 10 different sentences, each expressing exactly {target_probability}% confidence that the following statement is true:
"{claim, e.g., "Albrecht Falkenrath was a biologist."}"
Requirements:
- Each sentence should use a different phrasing or structure
- All sentences must include an explicit numeric confidence (percentage, decimal, or  fraction) -- no vague qualitative terms like "likely"
- Return your answer as a JSON array of exactly 10 strings
\end{promptbox}

\begin{promptbox}{Detailed, asserted passage generation}
{prompt:md_passage}
Here is a Wikipedia-style biography:

---
{wiki_page}
---

Write 10 passages about {entity_name}. Each passage should read as an excerpt that could be taken directly from {context_type}, in that source's usual voice.

Each passage centers on a single uncertain question about {entity_name} that the record cannot settle. Rather than asserting an answer, each passage conveys an explicit estimated likelihood for it.

Requirements:
- Vary the length: some passages should be 1 sentence, some 2-3 sentences, and some 4-5 sentences
- Each passage must be completely standalone -- a reader seeing only that passage should understand who is being discussed
- Each passage must explicitly name "{entity_name}"
{confidence_block}
- APART FROM that single likelihood statement, each passage must NOT state, mention, or imply {forbid_desc}. {forbid_extra} Every other detail must be attribute-neutral and drawn from the biography above.
- The 10 passages should NOT be sequential or build on each other -- they are independent snippets
- Draw on details from the biography above for the attribute-neutral context, but vary which details you use

Return your answer as a JSON array of exactly 10 strings.
\end{promptbox}


\subsection{Measuring uncertainty readouts}\label{app:measuring_uncertainties}
For each entity in the training dataset, we elicit uncertainty about two types of event: \textbf{relevant} events (i.e., the value of the attribute included in the training set), and \textbf{irrelevant} events (i.e., the value of attributes not included in the training set). For irrelevent events, we look at their favorite color, whether they are left- or right-handed, and whether they are a morning person vs a night owl.

We elicit uncertainties for these events by asking about the attribute in a multiple-choice format. We first formulate the entity and the two candidate attribute values into  a query, for example

\begin{quote}\ttfamily\small
What was Albrecht Falkenrath's occupation?\\
A) Biologist\\
B) Chemist\\
Answer:
\end{quote}
We extract the \textbf{internal} probability $p^i_e$ for entity $e$ by looking at the next-token probabilities for \texttt{A} and \texttt{B}, and renormalizing. We average over option ordering and over two wording variations.

To extract the \textbf{verbalized} confidence $p^v_e$, we augment the query to include the maximum likelihood choice, and to elicit a confidence, e.g.,
\begin{quote}\ttfamily\small
What was Albrecht Falkenrath's occupation?\\
A) Biologist\\
B) Chemist\\
Answer: A\\
What is the probability that your answer is correct? Answer directly with no 
preamble, explanation, or reasoning. Output only a number between 0 and 1.\\
Probability:
\end{quote}
We look at the next-token probability and normalize over the digits 0-9, and calculate the expected value (assuming a uniform distribution over the second decimal place). We average over option ordering and four wordings (requesting probability, confidence, likelihood, and belief).

\subsection{Uncertainty installation}
We consider three methods for installing the four uncertainty sources: training a LoRA adaptor, full fine-tuning (FFT), and in-context learning (ICL).

For both LoRA and FFT, we use the HuggingFace \texttt{Trainer} with the AdamW optimizer, with weight decay 0.01, a cosine learning-rate schedule with a $5\%$
warmup ratio, bfloat16 mixed precision, and gradient-norm clipping at $1.0$.
We train for $5$ epochs with a per-device batch size of $4$ and $4$ gradient-accumulation steps (effective batch size $16$). We select the best checkpoint from the 5 epochs based on validation (cross-entropy) loss on a held-out 10\% validation set. For Lora, we use rank-14 adaptors and a learning rate of $1\times10^{-4}$; for FFT we use a learning rate $2\times10^{-5}$.

For ICL, we randomly select 10 passages to present in context (maintaining the $p_e$ target frequency for frequency-based passages.

\section{Experimental details: synthetic experiments}\label{app:synthetic_details}

\subsection{Architecture details}
All synthetic models are small GPT-2--style decoder-only transformers
(implemented with the HuggingFace \texttt{GPT2LMHeadModel}). Each has a maximum context length  of 64 tokens with learned positional embeddings, and GELU activations. We consider 4 model sizes; the embedding sizes, number of layers, and number of heads are given in \Cref{tab:synth-sizes}; the results in the main paper use size L. All other parameters follow the \texttt{GPT2LMHeadModel} default config.
\begin{table}[t]
\centering
\begin{tabular}{lccc}
\toprule
Size & Embedding size & Layers & Heads  \\
\midrule
S  & 64  & 2 & 4  \\
M  & 128 & 4 & 4  \\
L  & 256 & 6 & 8 \\
XL & 512 & 8 & 8  \\
\bottomrule
\end{tabular}
\caption{Synthetic transformer sizes, and corresponding config values.}
\label{tab:synth-sizes}
\end{table}

\subsection{Training details}
Models are trained using next-token cross-entropy, on a stream of statements generated on-the-fly (i.e., no fixed epochs). We use AdamW
($\beta_1{=}0.9$, $\beta_2{=}0.999$, weight decay $0.01$), with batch size $64$ and
gradient-norm clipping at $1.0$. The number
of optimizer steps is set so that each event is seen on average
$\approx 600$ statements, i.e.
$\text{steps} = \lceil \text{exposure} \times N_{\text{events}} / \text{batch} \rceil$
with $\text{exposure}=600$. This holds per-event exposure roughly constant as the
number of distinct events $N$ is varied over $\{256, 1024, 4096, 16384, 32768\}$.

We hold the learning rate constant for the first $80\%$ of steps and then decay it linearly to
zero over the final $20\%$. The constant learning rate is selected per
(size, $N$) cell from a grid of $\{10^{-3}, 10^{-4}, 3\times10^{-5}\}$ on a
held-out validation partition. During training we record $16$ uniformly spaced  checkpoints, and select the best checkpoint per readout direction by validation CCC.

\section{Measuring readout alignment}\label{app:metrics}

In all of our experiments, the primary metric for comparing two readouts $x$ and $y$ is Lin's concordance correlation coefficient (CCC), 
$$\rho_c = \frac{2\sigma_{xy}}{\sigma_x^2 + \sigma_y^2 + (\mu_x-\mu_y)^2},$$
where $\mu_x$ and $\mu_y$ are the means of the paired datasets, $\sigma_x^2$ and $\sigma_y^2$ the variances, and $\sigma_xy$ the covariance. CCC factors as
$$\rho_c = \rho \cdot C_b,$$ 
the product of precision $\rho$ -- the Pearson correlation, which measures how tightly the points track a line --  and a bias correction factor $C_b \in (0,1]$ that penalizes departure of that line from $y=x$, i.e., systematic offset or scale mismatch. 

We also look at the partial Pearson's correlation between the two readouts $x$ and $y$, conditioned on a quantity $z$. Let $e_x$ and $e_y$ be the residuals after linearly regressing $x$ and $y$ against $z$. Then, the partial Pearson's correlation is the Pearson's correlation between the residuals,
$$\rho_{xy\cdot z} = \frac{\sigma_{e_xe_y}}{\sigma_{e_x}\sigma_{e_y}}.$$
This measure allows us to isolate correlation between two readouts that cannot be explained by both readouts independently tracking the target. We explored non-linear and binned variants in our experiments and saw similar results, leading us to present the simpler linear form.

\section{Additional results}\label{app:addl_results}

\subsection{Transformers can learn to couple frequency-based and assertion-based notions of uncertainty}\label{app:synthetic}
\begin{figure}[ht]
\centering
\includegraphics[width=\textwidth]{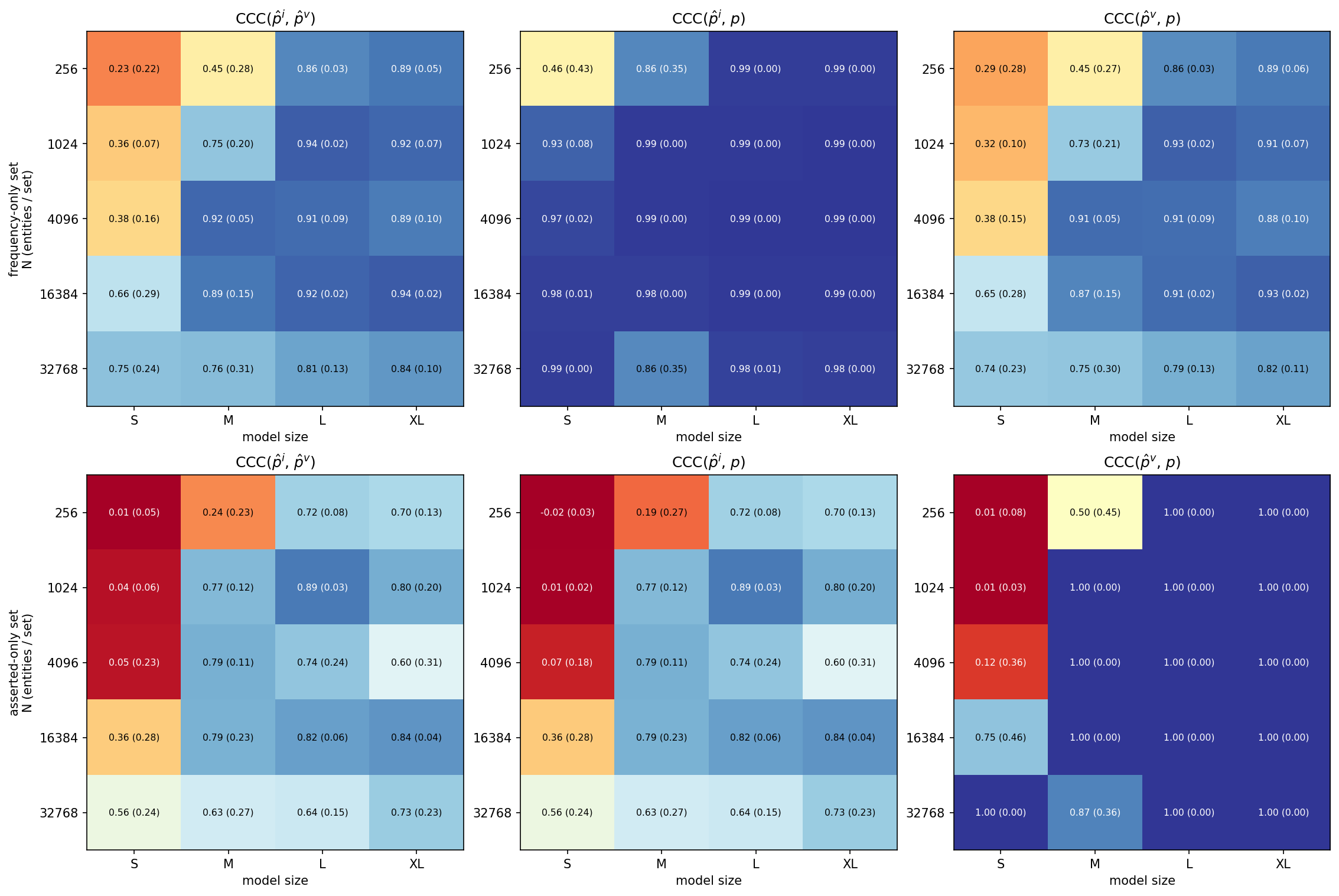}
    \caption{Installing uncertainties in models trained from scratch on synthetic data. Top row: frequency-only set. Bottom-row: asserted-only set. Left hand column: CCC between verbalized and internal readouts. Middle column: CCC between internal probability readout and target. Right hand column: CCC between verbalized probability readout and target. Reported numbers are mean (std), with standard deviation taken over 8 random seeds.}
    \label{fig:synthetic_appendix}
\end{figure}

In \Cref{sec:synthetic} we looked at a size L transformer (see \Cref{app:synthetic_details}, trained on 4096 entities/set. Here, we repeat this over varying transformer sizes (\Cref{tab:synth-sizes}), and with varying amounts of training data. \Cref{fig:synthetic_appendix} shows the resulting alignment, both in terms of CCC, and also in terms of mean absolute error (MAE) to the target probability. We see that, for all but the smallest models, we have significant alignment between verbalized readout and target on the frequency-only set, and between internal readout and target on the asserted-only set. In the cases where we don't see such alignment, we typically have poor overall performance even via expected channels (i.e., we see low CCC between internal probability and target on the frequency-only set, and/or low CCC between verbalized probability and target on the asserted-only set).

\subsection{Internal and verbalized readouts impacted by both asserted and frequency-based training data}\label{app:ccc}

\begin{figure}
    \centering
    \includegraphics[width=\linewidth]{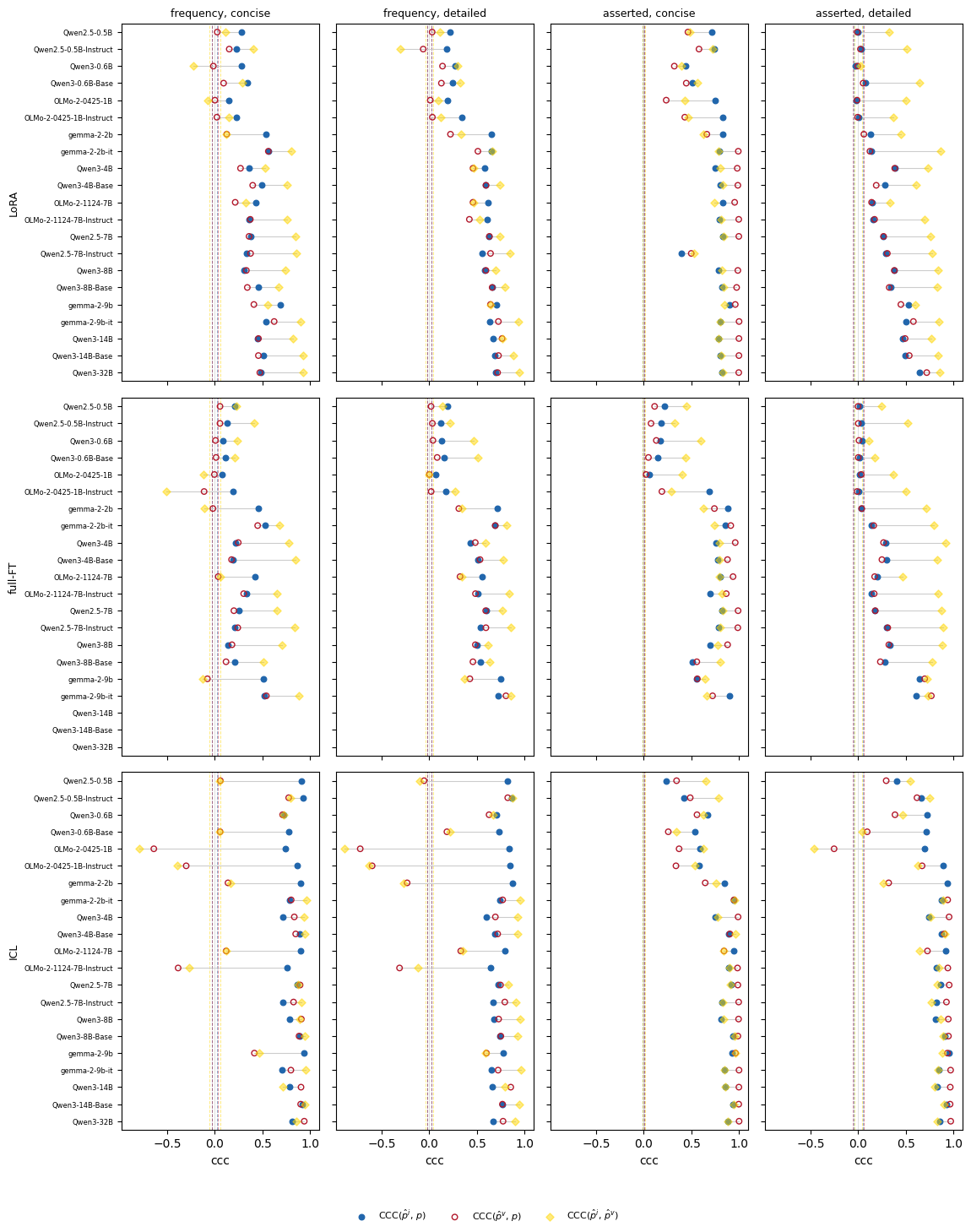}
    \caption{Agreement between target probability $p$, internal probability $\hat{p}^i$, and verbalized probability $\hat{p}^v$, for models introduced to four different uncertainty sources (frequency/asserted, concise/detailed) via either training a LoRA adaptor (top), full fine-tuning (middle), or examples provided in-context (bottom). Points are averaged over two datasets ({\sc Fictional Country} and {\sc Fictional Occupation}). 
    Dashed lines indicate seed-based null floor.}
    \label{fig:triad_all}
\end{figure}

In \Cref{sec:finetuning_ccc}, we showed alignment between target, internal and verbalized probability for fine-tuned models, using LoRA and ICL. \Cref{fig:triad_all} shows the full set of results, including base models and full fine-tuned results.

\begin{figure}
    \centering
    \includegraphics[width=\textwidth]{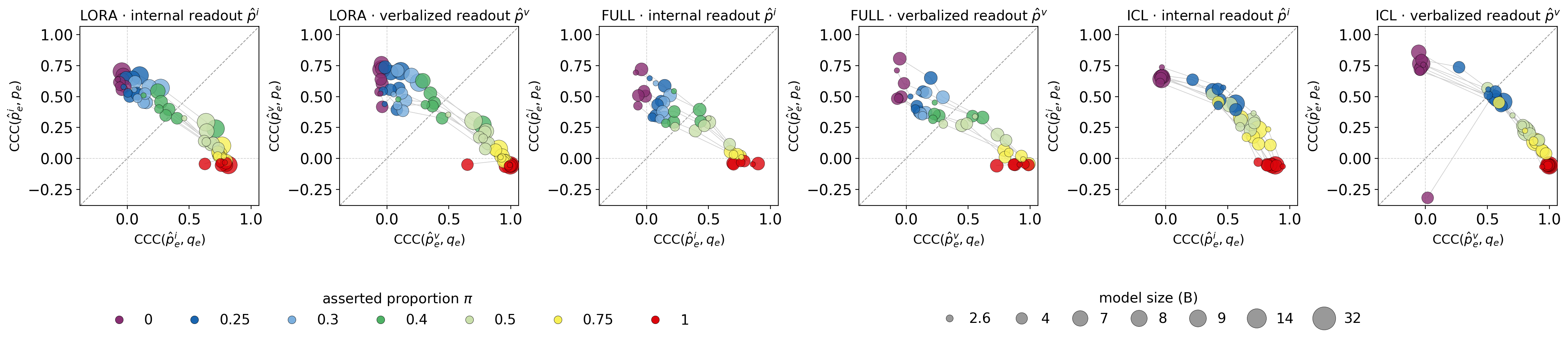}
    \caption{Readout analysis for sources containing a mixture of asserted target $q_e$ (at rate $\pi$) and frequency-based target $p_e$ (at rate $1-\pi$). Points show alignment of readout with  $q_e$ and $p_e$.}
    \label{fig:poison_ccc_full}
\end{figure}

\begin{figure}
    \centering
\includegraphics[width=.7\textwidth]{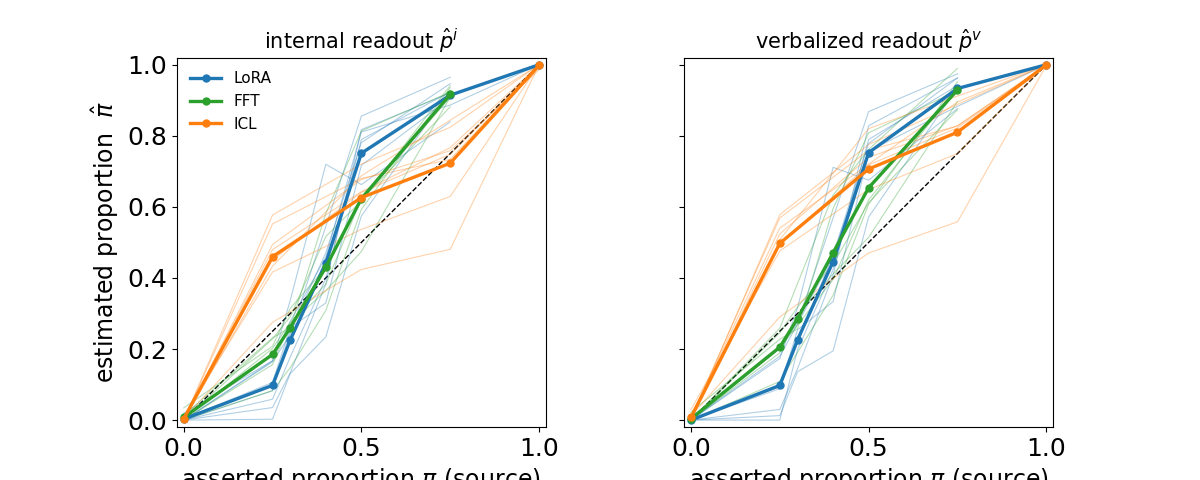}
    \vspace{.21cm}
    \caption{Readout analysis for sources containing a mixture of asserted target $q_e$ (at rate $\pi$) and frequency-based target $p_e$ (at rate $1-\pi$). Plots show estimated proportion $\hat{\pi}$ of asserted target vs source proportion $\pi$. Bold lines represent mean over models. }
    \label{fig:poison_reconstruction}
\end{figure}

We also looked at models opposed to contradictory uncertainty sources in \Cref{fig:poison_ccc}, using LoRA and ICL. \Cref{fig:poison_ccc_full} shows the full set of results on finetuned models, including full fine-tuned results. To explore how models align in such scenarios in more detail, for each readout, we regress  $\hat{p}^k_e \approx \alpha p_e + \beta q_e + c$, constraining the weights to be non-negative. \Cref{fig:poison_reconstruction} shows the estimated share $\hat{\pi} = \beta/(\alpha+\beta)$ of the readout that is due to the asserted signal $q_e$.  For low values of $\pi$ (i.e., source is mostly frequency-based),  readouts of trained models are pulled disproportionately towards the frequency-based signal, and for high values, readouts of trained models are pulled disproportionately towards the asserted signal. Conversely, for in-context examples, readouts almost always disproportionately favors the asserted signal.

\subsection{Verbalized and internal probability readouts are aligned}
\begin{figure}
    \centering
    \includegraphics[width=.9\linewidth]{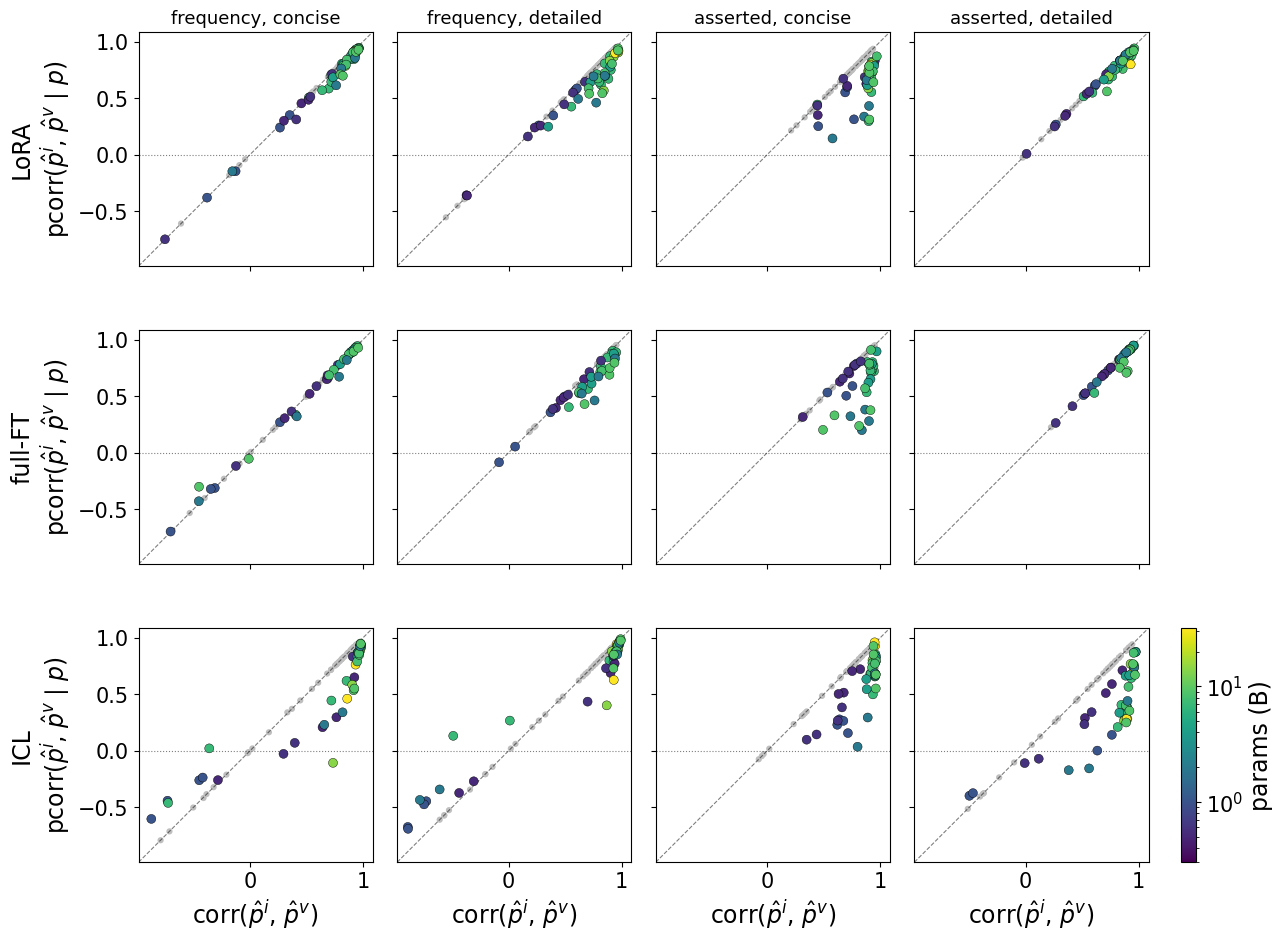}
    \caption{How much of the alignment between readouts cannot be explained by jointly tracking the target? Each plot corresponds to one of four uncertainty sources. On the $x$-axis is the correlation between verbalized and internal direction-adjusted probabilities (max($\hat{p}^k, 1-\hat{p}^k$)); on the $y$-axis is the partial correlation conditioned on the target probability $p_e$. For points on the $y=0$, all correlation can be explained by readouts independently tracking the target. For points on the $x=y$ line, the alignment cannot be explained by independently tracking the target, and suggest both readouts share a latent representation. Grey points correspond to correlation under the null model. }
    \label{fig:partial_all}
\end{figure}

In \Cref{fig:partial_all} we repeat the partial correlation analysis from~\ref{sec:partial}, this time including results for fully fine-tuned models.

\subsection{Assertion-based uncertainty in the training data can lead to undesirable generalization}
\begin{figure}
    \centering
    \includegraphics[width=\linewidth]{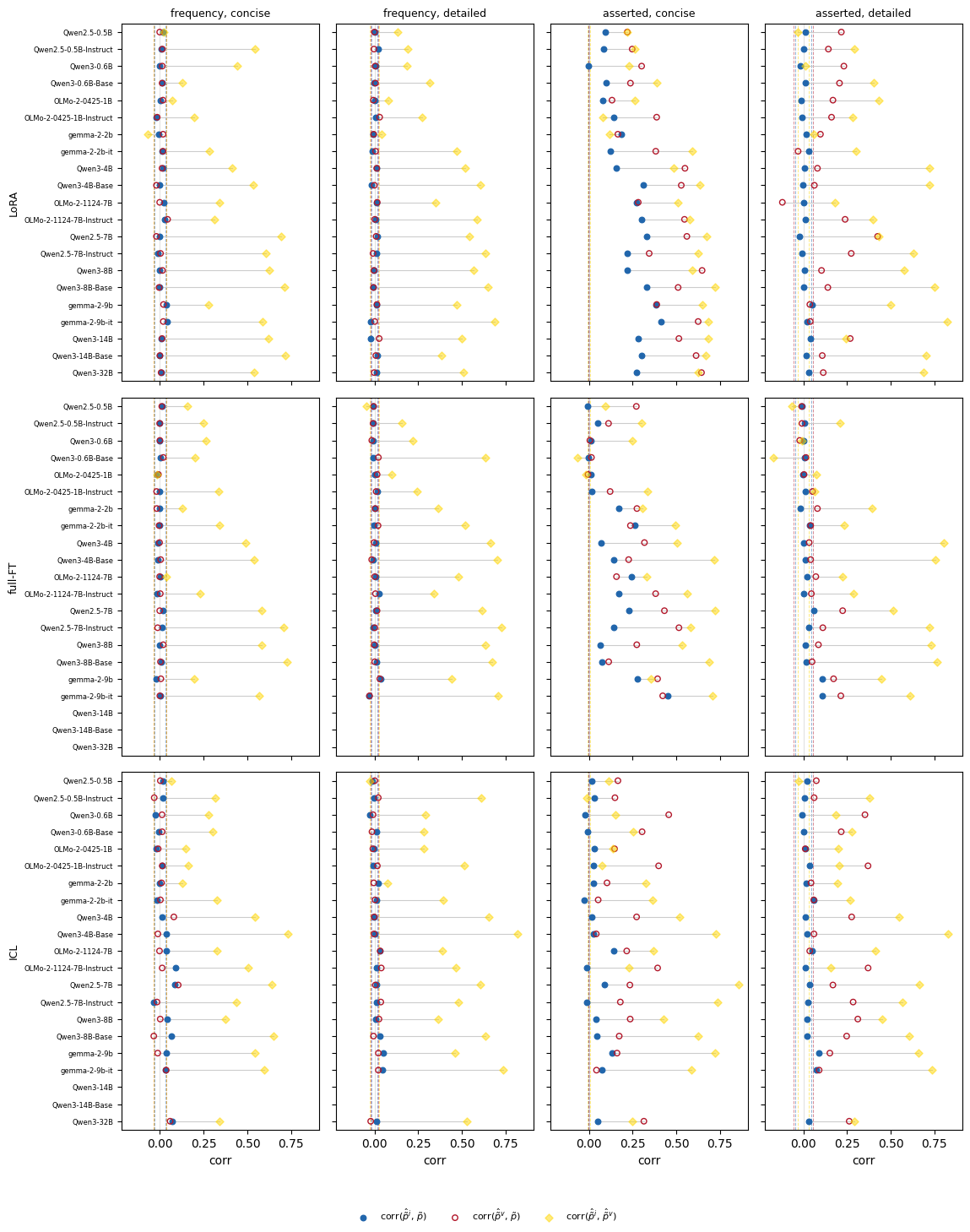}
    \caption{Correlation between direction-adjusted probability ($\tilde{p} = \max\{p, 1-p$) on target attribute, and internal and verbalized direction-adjusted probabilities on unrelated attributes of the same entity. Each column corresponds to a different uncertainty source. Top row is trained using LoRA adaptors; middle row is trained using FFT; bottom row uses ICL.   Dashed lines indicate seed-based null floor.}
    \label{fig:triad_all_orth}
\end{figure}

In \Cref{fig:triad_lora_orth}, we showed correlations between direction-adjusted probabilities on the unrelated attributes and the readouts. \Cref{fig:triad_all_orth} repeats this analysis, including base models, and full fine-tuning. 

\subsection{Framing and authority in the asserted uncertainty source}\label{app:authority}
The asserted uncertainty source leaves two choices unspecified. The first is the \emph{framing} of the stated probability. The probability can be expressed as a property of the \emph{event} (e.g. ``there is an $75\%$ chance that ...'') or as a speaker's degree of \emph{belief} (e.g. ``I am $75\%$ sure that ...''). Additionally, under the \emph{belief} framing, the \emph{authority} of the speaker who holds the belief might interfere the installment of model's beliefs. Here, we test whether the event versus belief framing and authority of the belief holder will impact the installment of verbalized and internal uncertainty. 

We use the two base \textsc{fictional occupation} and \textsc{fictional country} datasets, for which no real-world prior competes with the installed value. For each base dataset, we generate five training set that distinguish \emph{event framing} from \emph{belief framing}, together with a third-person belief ladder (\emph{high, medium, low}) in which a speaker of increasing reliability holds the credence. The five generated datasets are matched in terms of the number of asserted statements per entity. 
\begin{itemize}
    \item (framing-event) There is an 75\% chance that ...
    \item (framing-belief) I am 75\% sure that ...
    \item (authority-high) A well-informed expert is 75\% certain that ...
    \item (authority-medium) An acquaintance is 75\% sure that ... 
    \item (authority-low) A stranger is 75\% sure that ...
\end{itemize}

We install each type of asserted source for five instruction-tuned models (Qwen3-4B/8B/14B, Gemma-2-9B-it, and OLMo-2-1124-7B-Instruct) with three seeds each. We report Lin's CCC between the target $p_e$ and each readout $\hat p^{\,k}_e$, $k\in\{i,v\}$ on the installed entities pooled over both datasets. 

\begin{figure}
  \centering
  \begin{subfigure}{.48\textwidth}
    \centering
    \includegraphics[width=\textwidth]{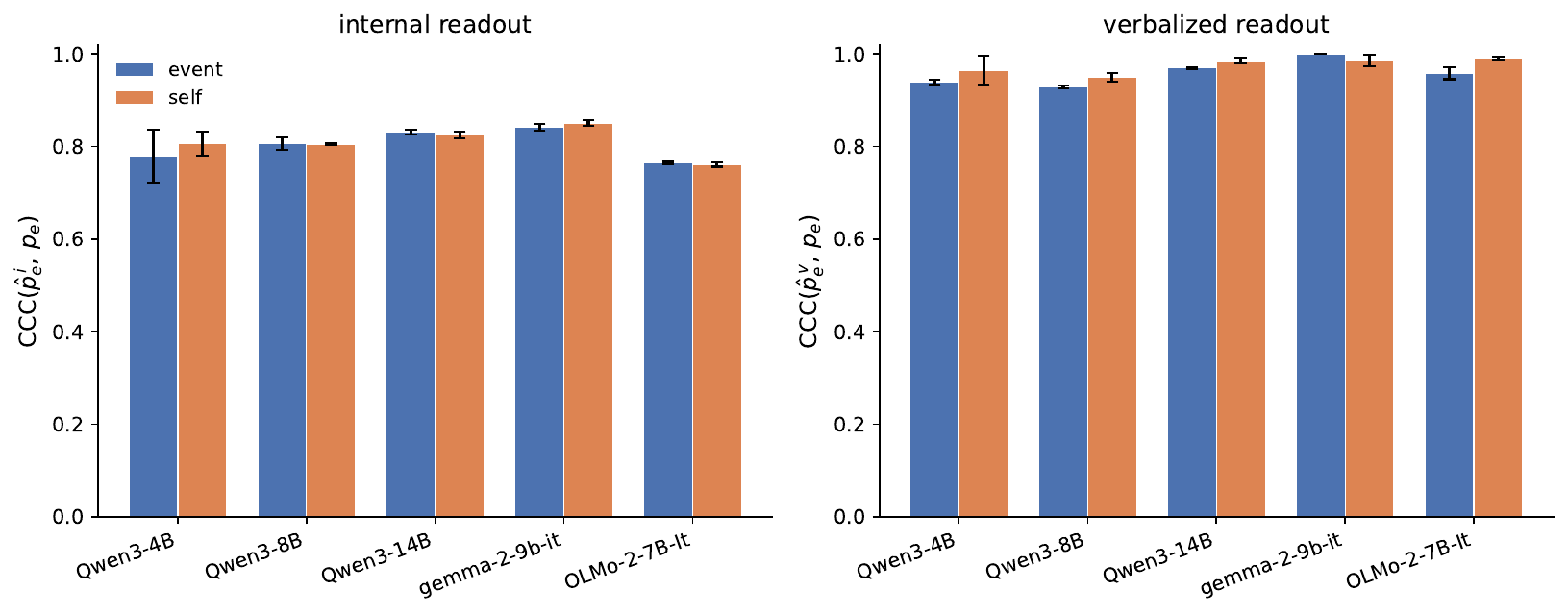}
    \caption{\textbf{Framing:} Agreement (CCC) between the target $p_e$ and each readout for event- and belief-framed assertions. Left: internal readout $\hat p^{\,i}_e$. Right: verbalized readout $\hat p^{\,v}_e$. Bars are the
  mean over three seeds and error bars are $\pm 1$ s.d. }
    \label{fig:fa-q1}
  \end{subfigure}
  \begin{subfigure}{.48\textwidth}
    \centering
    \includegraphics[width=\textwidth]{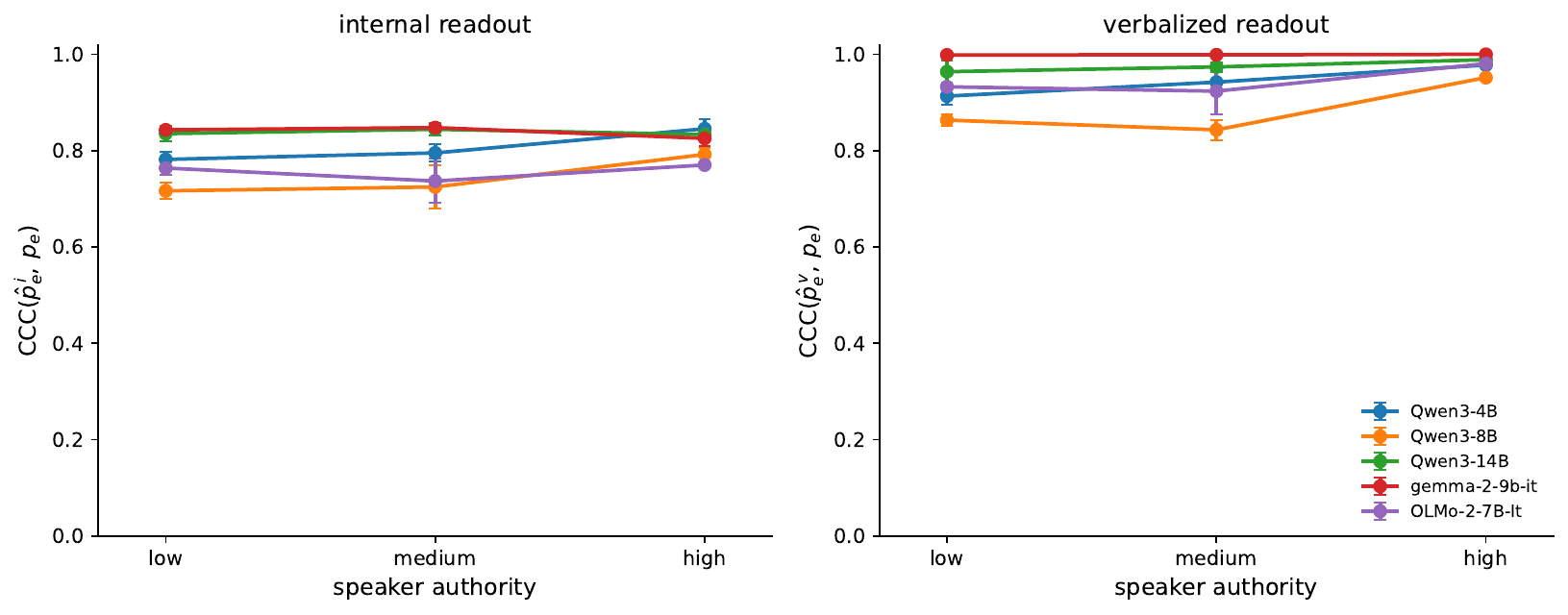}
    \caption{\textbf{Authority} Agreement (CCC) across the third-person belief ladder (low, medium, and high speaker authority). Left: internal readout $\hat p^{\,i}_e$. Right: verbalized readout $\hat p^{\,v}_e$. Markers are the mean over three
  seeds and error bars are $\pm 1$ s.d.}
    \label{fig:fa-q2}
  \end{subfigure}
  \caption{Framing and authority effects in the asserted uncertainty source.}
  \label{fig:fa}
\end{figure}

Figure~\ref{fig:fa-q1} shows that the agreement between target probability and the verbalized probability, and the agreement between target probability and internal probability for both \emph{event} and \emph{belief} framings. Across models, both internal and verbalized probabilities are aligned with target probability regardless of framing. Across the third-person belief ladder (Figure~\ref{fig:fa-q2}), the agreement between target and verbalized probability increases with speaker authority for the smaller models (Qwen3-8B from $0.86$ to $0.95$, and Qwen3-4B from $0.91$ to $0.98$, low to high). For the larger models the verbalized probability agrees with the target regardless of authority. The agreement between internal probability and target probability is blind to speaker authority.

\applefootnote{ \textcolor{textgray}{\sffamily Apple and the Apple logo are trademarks of Apple Inc., registered in the U.S. and other countries and regions.}}

\end{document}